\documentclass{article}

\usepackage{microtype, xspace}
\usepackage{tcolorbox}
\usepackage{graphicx}
\usepackage{multirow} 

\usepackage{booktabs}
\usepackage{graphicx, array} %
\usepackage{amssymb}%

\usepackage{hyperref}

\usepackage[accepted]{icml2025}

\usepackage{amsmath}
\usepackage{amssymb}
\usepackage{mathtools}
\usepackage{amsthm}
\usepackage{comment}
\usepackage{enumerate, enumitem}
\usepackage{subcaption}

\usepackage[capitalize,noabbrev]{cleveref}

\theoremstyle{plain}
\newtheorem{theorem}{Theorem}[section]

\theoremstyle{definition}

\theoremstyle{remark}
\newtheorem{remark}[theorem]{Remark}

\def\cA{{\mathcal{A}}} \def\cB{{\mathcal{B}}}  \def\cD{{\mathcal{D}}}
   \def\cH{{\mathcal{H}}}
   \def\cL{{\mathcal{L}}}
   
 \def\cR{{\mathcal{R}}} \def\cS{{\mathcal{S}}}

\def\ba{{\mathbf{a}}}    
    
\def\bk{{\mathbf{k}}}    
\def\bp{{\mathbf{p}}} \def\bq{{\mathbf{q}}} \def\br{{\mathbf{r}}} \def\bs{{\mathbf{s}}} 
    
 \def\bh{\mathbf{h}}

 \def\bQ{{\mathbf{Q}}}

\def\lang{\texttt{lang}} 

\def\task{\texttt{task}\xspace}

\def\llm{\textrm{LLM}}

\def\arule{\ba^{\text{rule}}}
\def\aenv{\ba^{\text{env}}}
\def\aexpl{\pmb{\ell}^{\text{expl}}}
\def\athought{\ba^{\text{thought}}}
\def\renv{\br^{\text{env}}}
\def\rrule{\br^{\text{rule}}}
\def\Rrule{R^{\text{rule}}}

\newcommand{\rbrl}{\texttt{RBRL}\xspace}

\newcommand{\rev}[1]{{#1}}

\usepackage[textsize=tiny]{todonotes}

\icmltitlerunning{RBRL: Joint Explanation and Decision Optimization for Resource Allocation with Language Agents}

\begin{document}

\twocolumn[

\icmltitle{
    Rule-Bottleneck Reinforcement Learning: Joint Explanation and Decision Optimization for Resource Allocation with Language Agents%
}

\icmlsetsymbol{equal}{*}

\begin{icmlauthorlist}
\icmlauthor{Mauricio Tec}{equal,seas,hsph}
\icmlauthor{Guojun Xiong}{equal,seas}
\icmlauthor{Haichuan Wang}{seas}
\icmlauthor{Francesca Dominici}{hsph}
\icmlauthor{Milind Tambe}{seas,deepmind}
\end{icmlauthorlist}

\icmlaffiliation{seas}{Department of Computer Science, Harvard John A. Paulson School of Engineering and Applied Sciences}
\icmlaffiliation{hsph}{Department of Biostatistics, Harvard T.H. Chan School of Public Health}
\icmlaffiliation{deepmind}{Google DeepMind}

\icmlcorrespondingauthor{Mauricio Tec}{mauriciogtec@g.harvard.edu}
\icmlcorrespondingauthor{Guojun Xiong}{gjxiong@g.harvard.edu}

\icmlkeywords{Machine Learning, ICML}

\vskip 0.3in
]

\printAffiliationsAndNotice{\icmlEqualContribution} %

\begin{abstract}
Deep Reinforcement Learning (RL) is remarkably effective in addressing sequential resource allocation problems in domains such as healthcare, public policy, and resource management. However, deep RL policies often lack transparency and adaptability, challenging their deployment alongside human decision-makers. In contrast, Language Agents, powered by large language models (LLMs), provide human-understandable reasoning but may struggle with effective decision making. To bridge this gap, we propose Rule-Bottleneck Reinforcement Learning (\rbrl), a novel framework that jointly optimizes decision and explanations. At each step, \rbrl generates candidate rules with an LLM, selects among them using an attention-based RL policy, and determines the environment action with an explanation via chain-of-thought reasoning. The RL rule selection is optimized using the environment rewards and an explainability metric judged by the LLM. Evaluations in real-world scenarios highlight \rbrl's competitive performance with deep RL and efficiency gains over LLM fine-tuning. A survey further confirms the enhanced quality of its explanations.
\end{abstract}

\section{Introduction}

Sequential resource allocation is a fundamental problem in many domains, including healthcare, finance, and public policy \cite{considine2023optimizing,boehmer2024optimizing, yu2024fincon}. This task involves allocating limited resources over time while accounting for dynamic changes and competing demands. Deep reinforcement learning (RL) is an effective method to optimize decision-making for such challenges, offering efficient and scalable policies~\cite{yu2021reinforcement,talaat2022effective, xiong2023reinforcement,zhao2024towards}. However, deep RL policies generally provide action recommendations without human-readable reasoning and explanations. Such lack of interpretability poses a major challenge in critical domains where decisions must be transparent, justifiable, and in line with human decision-makers to ensure trust and compliance with ethical and regulatory standards.

For example, doctors may need to decide whether to prioritize intervention for Patient A or Patient B based on their current vital signs~\cite{boehmer2024optimizing}. An RL algorithm might suggest: \textit{ ``Intervene with Patient A "} with the implicit goal of maximizing the value function. However, the underlying reasoning may not be clear to the doctors, leaving them uncertain about the factors influencing the decision \cite{milani2024explainable}. For doctors, a more effective suggestion could be risk-based with specific information, e.g., \textit{``Patient A's vital signs are likely to deteriorate leading to higher potential risk compared to Patient B, so intervention with Patient A is prioritized"} \cite{gebrael2023enhancing, boatin2021wireless}.

\begin{figure*}[tbp]
    \centering
    \includegraphics[width=0.99\linewidth]{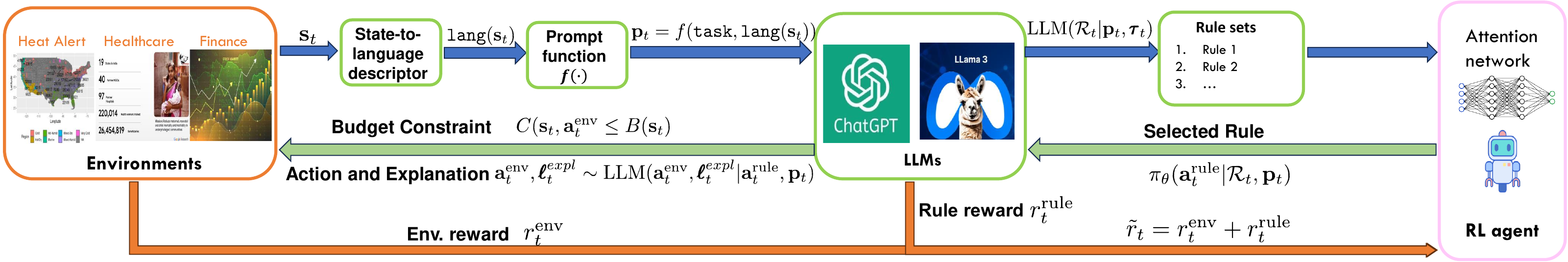}
    \caption{Overview of the \rbrl framework for joint sequential decision-making and explanation generation at time instance $t$. Starting with current state $\bs_t$,  a state-to-language descriptor generates \lang($\bs_t$), which is used to create the input prompt 
$\bp_t$. The LLM processes 
$\bp_t$
  to produce a thought 
$\pmb{\tau}_t$  and a set of candidate rules 
$\cR_t$ . An attention-based policy network selects a rule 
$\arule_t$ , which is used to derive an executable action $\aenv_t$ satisfying the budget constraint $B(\bs_t)$ for the environment 
  and a human-readable explanation $\pmb{\ell}_t^{expl}$, while also providing a rule reward $r_t^{\text{rule}}$ 
 . The environment transitions to the next state 
$\bs_{t+1}$ , returning an environment reward $r_t^{\text{env}}$ 
 . This process is repeated iteratively at subsequent time steps. 
}
    \label{fig:Proposed_framework}
\end{figure*}

Language agents \cite{sumers2024cognitive} leverage large language models (LLMs) for multi-step decision-making using reasoning techniques like chain of thought (CoT) \cite{wei2022chain} and ReAct \cite{yao2023react}. They enable natural language goal specification \cite{du2023guiding} and enhance human understanding \cite{hu2023language, srivastava2024policy}. However, LLMs struggle with complex sequential decision-making, such as resource allocation \cite{furuta2024exposing}, making RL a crucial tool for refining them into effective policy networks \cite{carta2023grounding, tantrue, wen2024reinforcing, zhai2024fine}. Yet, fine-tuning LLMs for policy learning is highly challenging due to the substantial computational costs and the complexity of token-level optimization \cite{rashid2024critical}, which remains an open challenge, particularly in sequential resource allocation.

Consequently, aiming to combine the strengths of both deep RL and language agents, we pose the following question:

\vspace{-0.1in}
\begin{tcolorbox}[colback=white!5!white,colframe=white!75!white]
\textit{%
Can we design a language agent framework that can simultaneously perform sequential resource allocation and provide human-readable explanations? }
\end{tcolorbox}
\vspace{-0.15in}

Motivated by existing work that employs predefined rules or concepts to explain RL policies \cite{Das2023State2Explanation} or guide RL exploration \cite{likmeta2020combining}, we explore the potential of using rules to prioritize individuals in resource allocation problems. In the context of language agents, rules are defined as ``structured statements" that capture prioritization among choices in a given state, aligning with the agent's goals \cite{srivastava2024policy}. 
Rules offer a flexible framework for encoding high-level decision criteria and priority logic, similar as the celebrated index policy for prioritizing arms in resource allocation problems \cite{whittle1988restless}, making them ideal for guiding resource allocation strategies while explaining the rationale behind decisions.%

Building on this, we propose a novel framework called Rule-Bottleneck Reinforcement Learning (\rbrl), which integrates the strengths of LLMs and RL to bridge the gap between decision-making and interoperability, by optimizing LLM-generated rules with RL. 
\rbrl provides a framework (as shown in Figure \ref{fig:Proposed_framework}) that simultaneously makes sequential resource allocation decisions and provides human-readable explanations. \rbrl leverages LLMs to generate candidate rules and employs RL to optimize policy selection, enabling the creation of effective decision policies while simultaneously providing human-understandable explanations. 

Our contributions are summarized as follows. \textit{First}, to avoid the computational cost and complexity of directly fine-tuning language agents, we leverage LLMs to generate a diverse set of rules, where each rule serves as a prioritization strategy for individuals in resource allocation. This approach enhances flexibility and interpretability in decision-making.
\textit{Second}, we extend the conventional state-action space by integrating the thoughts and rules generated by LLMs, creating a novel framework that enables reinforcement learning to operate on a richer, more interpretable decision structure.
\textit{Third}, we introduce an attention-based training framework that maps states to queries and rules to keys. The rule selection process is optimized by a policy network trained using the Soft Actor-Critic (SAC) algorithm \cite{haarnoja2018soft}, ensuring robust and efficient decision-making. In particular, the LLM also acts as a feedback mechanism, providing guidance during RL exploration to improve policy optimization and promote more effective learning.

We evaluate our method in three environments from two real-world domains: \texttt{HeatAlerts}, where resources are allocated to mitigate extreme heat events; and \texttt{WearableDeviceAssignment}, for distributing monitoring devices to patients. 
Using cost-effective LLMs such as gpt-4o-mini \cite{openai2024gpt4omini} and Llama 3.1 8B \cite{meta2024llama3.1}, we first assess decision performance by comparing \rbrl with pure RL methods and language agent baselines. We then evaluate explanation quality through a human survey conducted under IRB approval. The results demonstrate \rbrl's effectiveness in both decision quality and interpretability.

\section{Related Work}
\label{sec:related}

Our work intersects with three distinct areas within the RL literature. We discuss related work in each of these domains.

\textbf{RL for Resource Allocation}\quad
RL has been widely studied for constrained resource allocation across domains. In maternal health, \citet{boehmer2024optimizing} apply RL to a restless multiarmed bandit (RMAB) problem \cite{whittle1988restless} to compute patient-specific intervention probabilities. Also in an RMAB setting, \citet{xiong2022index} propose a model-based RL approach that prioritizes users via an index and allocates resources under budget constraints. In public health, \citet{considine2023optimizing} propose RL to optimize  extreme heat warnings under a budget on the number of possible alerts. Other works include multi-agent RL for robotic warehouse allocation \cite{Shen2023} and exogenous MDPs for cloud resource management \cite{sinclair2023hindsight}. While these methods optimize rewards effectively, they often lack interpretability---critical for deployment in sensitive domains requiring trust, transparency, and accountability.

\textbf{RL and Language Agents}\quad 
The language agents \citep{sumers2024cognitive} paradigm developed somewhat independently of RL, with works like ReAct prompting \cite{yao2023react} extending chain-of-thought (CoT) \cite{wei2022chain} to action settings. These works have focused on tasks such as open-ended web navigation \citep{putta2024agent}, social simulations \citep{park2023generative}, and virtual assistants \citep{vezhnevets2023generative}. Meanwhile, language interfaces have also be been proposed within the RL literature, including leveraging external and commonsense knowledge \cite{feng2024natural, waytowich2024atari}, pre-training goal-based policies \cite{du2023guiding}, enhancing generalization in embodied agents \cite{szot2023large}, and aiding human-AI coordination \cite{srivastava2024policy, hu2023language}. Related works include GLAM \citep{carta2023grounding}, TWOSOME \cite{tantrue}, BAD \cite{wen2024reinforcing}, and TextGym \cite{xi2024agentgym}, which use LLM finetuning techniques in RL environments with a reward function. Relevant to our work is also \citet{wang2023describe}, which, inspired by open-ended settings like Minecraft, employs RL to optimize the goals of an LLM planner based on feasibility.

\textbf{Explainable RL (XRL)}\quad
Early XRL relied on methods like decision trees and concept-based explanations \citep{Das2023State2Explanation}, but these struggled with scalability in dynamic environments \citep{poeta2023concept}. Recent advances introduced large language models (LLMs) for post-hoc explanations, such as explaining decision paths from policy trees \citep{zhang2023understanding} or adding language descriptions to RL policies \citep{colas2022language}. However, these approaches focus on interpreting pre-existing policies rather than enabling LLMs to generate inherently explainable decisions, with challenges in aligning explanations to human reasoning \citep{singh2024rethinking}. By contrast, inherently (also known as intrinsically) interpretable policies  are those that have internal representation that allow explanations 
 \cite{peng2022inherently,milani2024explainable}. Our work sits this literature by using LLM reasoning traces as the basis for environment action selection.
 
 \rev{
    With various works acknowledging the trade-off between interpretability and performance, prioritizing interpretability appears to be crucial in practice for many critical applications \cite{rudin2019stop}: an approach that we subscribe to in this work. For example, in the clinical AI domain, physicians require transparency to validate recommendations and uphold ethical accountability, as mandated by regulatory frameworks (e.g., \citet{eu_ai_act_2024, ca_ai_act_2024}). High-performing black-box systems often face rejection in clinical workflows due to distrust \cite{shevtsova2024trust, dubois2019deep}.  By contrast, interpretable models allow clinicians to audit biases and adapt logic to local contexts, whereas opaque policies risk failures under real-world distribution shifts \cite{rudin2019stop,doshi2017towards}. Transparent reasoning facilitates iterative, clinician-driven refinement, ensuring collaborative decision aid rather than an inflexible oracle \cite{shevtsova2024trust, dubois2019deep}. Empirical surveys show clinicians favor models that enable shared decision-making, error accountability, and ethical oversight despite modest performance penalties \cite{shevtsova2024trust}—a critical stance in high-stakes healthcare environments where trust and adaptability outweigh narrow efficiency gains.
}

\section{Preliminaries and Problem Formulation}

\textit{Notations.} A bold lowercase letter $\ba \in\mathbb{R}^d$ represents a vector with dimension $d$, while a regular lowercase letter $a$ denotes a scalar. An uppercase letter $A$ denotes a mapping function, and a calligraphic letter $\cA$ denotes a set; $[h]$ denotes the set of $\{1, \ldots, h\}$; $\Delta(\cA)$ denotes the space of probability distributions supported in $\cA$.

\subsection{Resource-Constrained Allocation}
Resource-constrained allocation tasks are usually formulated as a special Constrained Markov Decision Process (CMDP), which is defined by the tuple $ \langle \cS, \cA, P, R, C, h,\gamma \rangle $, where $\cS$ denotes a state space and $\cA$ denotes a finite action space. The transition probability function, specifying the probability of transitioning to state $ \bs'\in\mathbb{R}^{d_1} $ after taking action $\ba\in\mathbb{R}^{d_2}$ in state $\bs$, is $ P(\bs' | \bs, \ba):\cS \times \cA \times \cS \to \Delta(\cS) $,  $ R(\bs, \ba):\cS \times \cA \to \mathbb{R}$ represents the reward function, defining the immediate reward received after taking action $\ba$ in state $\bs$, and we let $C(\bs, \ba): \cS \times \cA \to \mathbb{R}^{d_3}$ be the immediate cost incurred after taking action $\ba$ in state $\bs$. Often, each dimension $i\in[d_2]$ in $\ba$ is either $0$ or $1$ in resource-constrained allocation tasks. In addition, $h$ is the time horizon and $ \gamma \in [0, 1] $ denotes the discount factor, which determines the present value of future rewards.

The goal of resource-constrained allocation tasks is to find a policy $\pi\colon \cS \to \Delta(\cA)$ that maximizes the expected cumulative discounted reward while satisfying the cost constraints:
\begin{align}\label{eq:RCA_objective}
    \pi^* &= \arg\max_\pi \mathbb{E}_\pi J(\pi):=\left[ \sum_{t=1}^h \gamma^{t-1} R(\bs_t, \ba_t) \right]\nonumber\\
    & s.t. ~~~\forall t\in [h]\colon C(\bs_t, \ba_t) \leq B(\bs_t),
\end{align}
where $B\colon \cS \to \mathbb{R}^{d_3}$ is the budget function.

\begin{figure}[tbp]
    \centering
    \begin{subfigure}[b]{\linewidth}
\includegraphics[width=0.98\linewidth]{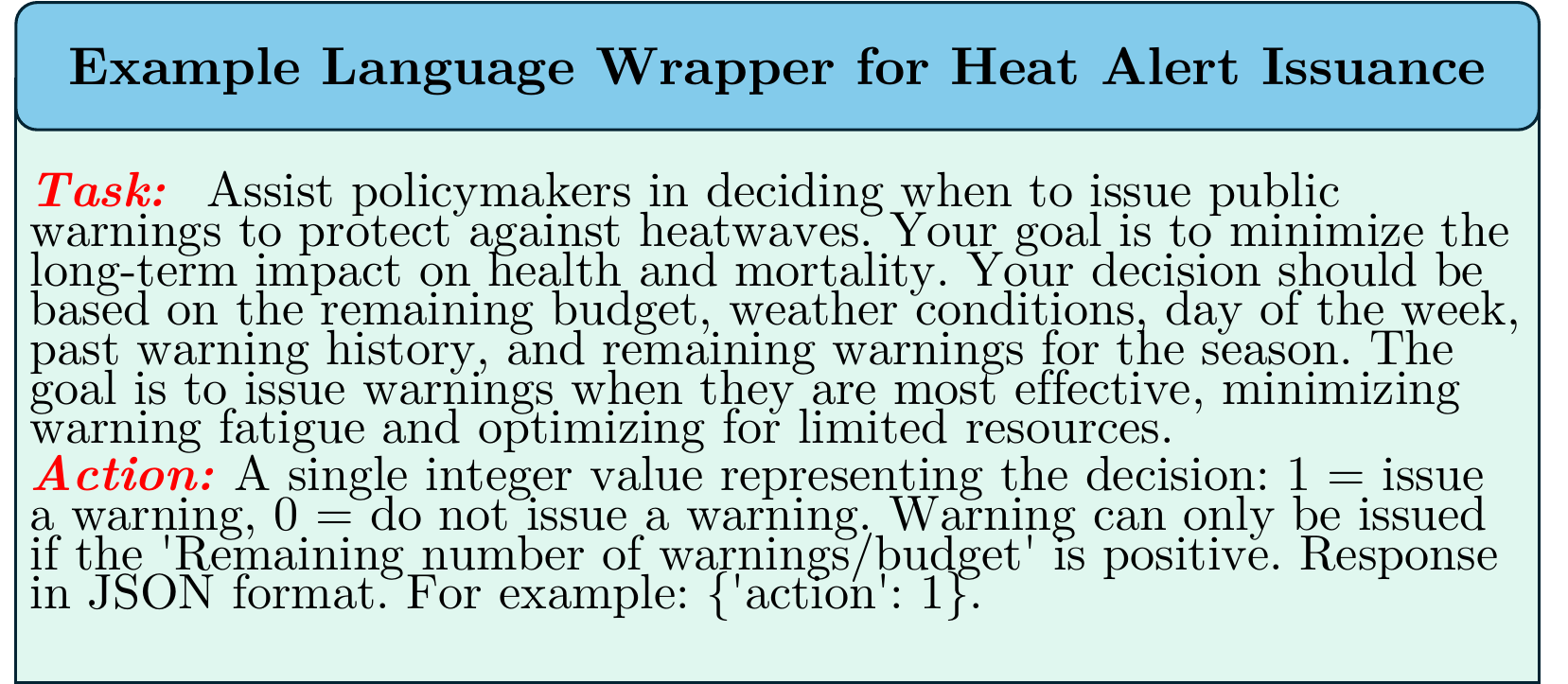}
    \caption{Examples of initial task prompt, which contains the task description and available actions. }
    \label{fig:input_prompt}   
    \end{subfigure}
     \begin{subfigure}[b]{\linewidth}\includegraphics[width=0.98\linewidth]{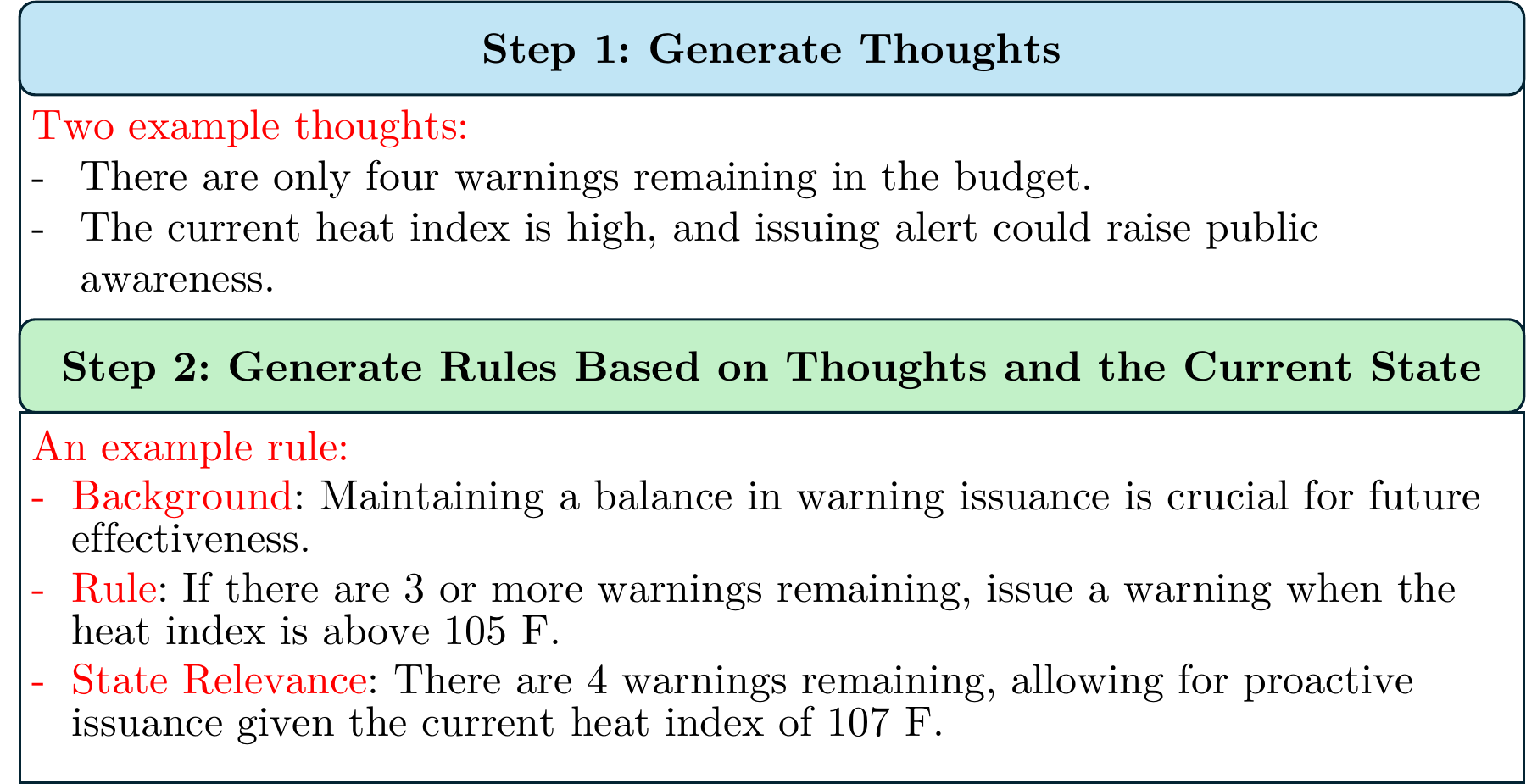}
    \caption{Examples of generated rules for the Heat Alert Issuance task.}
    \label{fig:rule_prompt}   
    \end{subfigure}
    \vskip -4pt
    \caption{Examples of task prompts and generated rules.}
    \vspace{-0.6cm}
\end{figure}

\subsection{Language-augmented RL with Rules}

We outline the language-augmented RL framework, where the action space includes internal language actions $\tilde{\cA}=\cA \cup \cL$ \cite{yao2023react, carta2023grounding}. Language agents typically have two types of internal language actions: First, \emph{thoughts} $\athought \in \cL$, are reasoning traces from the current problem state used to inform environment action selection $\aenv \in \cA$ \cite{yao2023react}. Second, \emph{post-hoc explanations} $\aexpl$, are generated from actions and thoughts to enhance human trust and interpretability \cite{zhang2023understanding}, a focus of this work.

\textbf{Rules}\quad
Thoughts are useful to highly relevant aspects of a problem. However, they often lack detailed information to identify the next optimal action. In this work, we will consider ``rules" $\arule\in \cL$, which are structured language statements derived from thoughts that generally take the form ``[do/prioritize] [if/when]". More formally, each rule $\arule$ consists of a triple $(\texttt{\footnotesize background}, \texttt{\footnotesize\, rule\_statement}, \texttt{\footnotesize\, state\_relevance})$. Figure \ref{fig:rule_prompt} shows examples of generated rules from one of the domains used in the experiments.

\textbf{Task and Constraints Description}\quad
Language agents require: (1) a language description of the environment and the agent's goal, denoted \task; (2) a function describing the state of the environment in natural language, denoted $\lang:\cS\to \cL$.  At each state $\bs_t$, these descriptors are used to construct a natural language prompt $\bp_t=f(\task, \lang(\bs_t), C, B)$, where $C$ and $B$ are the costs and budget constraints defined in eq. \eqref{eq:RCA_objective}. For instance, the prompt can restrict certain actions, or the total number of allocations in a given period of time.
Figure \ref{fig:input_prompt} exemplifies rules generated for the environments in our experiments.

\textbf{Baseline Rule-based Language Policy}\quad
Our objective is to jointly optimize the reward and explainability of the environment. Hence, we will take as baseline an LLM-driven policy $\pi_{\llm}$ for online interaction with the environment:
\begin{align}\label{eq:lang-policy}
\athought_t & \sim \pi_{\llm}(\athought_t \mid \bp_t), \allowdisplaybreaks\\\label{eq:lang-policy-rule}
\arule_t  & \sim \pi_{\llm}(\arule_t  \mid \athought_t, \bp_t), \\\label{eq:lang-policy-env}
\aenv_t   & \sim \pi_{\llm}(\aenv_t  \mid \arule_t, \athought_t, \bp_t),\allowdisplaybreaks \\ \label{eq:lang-policy-expl}
\aexpl_t  & \sim \pi_{\llm}(\aexpl_t \mid \aenv, \arule_t , \bp_t).
\end{align}

The rule acts as a ``bottleneck" to the action and explanation.

\begin{remark} This baseline builds on the baseline chain-of-though language agent, augmented by the generation of a single rule $\arule_t$. In the next section, we will introduce \rbrl, which replaces the generation of a single $\arule_t$ with an RL-based learnable selection policy $\pi_\theta$ choosing among a set of dynamically generated candidate rules.
\end{remark}

\textbf{Explainability}\quad
Explainability in AI encompasses multiple dimensions (see \citet{hoffman2018metrics} for an overview). Here, we evaluate the explainability of 
$\arule$ and $\aexpl$ through two key aspects:

\begin{enumerate}[leftmargin=2em,topsep=0pt,itemsep=0pt] \item \emph{Completeness of 
$\arule$}: Can the optimal action be predicted from the rule? Does it contain sufficient detail about its applicability to the current state?
\item \emph{Trust and understanding with 
$\aexpl$}: Do humans find the explanation useful to understand the system behavior? Does it foster trust in the system?
\end{enumerate}

Analyzing both $\arule$ and $\aexpl$ is critical since post hoc LLM explanations alone may be prone to ``satisfying" but misleading rationalizations \citep{zhang2023understanding}. We hypothesize that complete rules aid to generating better post-hoc explanations by outlining a more structured reasoning process.

\subsection{Problem Statement}

Our primary challenge is to enable LLMs to jointly optimize a language policy that both solves the underlying optimization problem and enhances the quality of the explanations---a task that has received little attention in the literature.  We aim to increase the quality of $\aexpl$ while also optimizing decision-making. We aim to achieve this by selecting rules that encourage both good quality explanations and high reward.  In Section \ref{sec:rbrl}, we will describe our method for constructing a surrogate explainabilty ``rule reward" \(\Rrule_\llm(\arule)\) using an LLM as judge  \citep{shen2024explainable, bhattacharjee2024towards, gu2024survey}. We denote the joint environment/rule reward as $ \tilde{r}_t = R(\bs_t, \aenv_t) + \Rrule_\llm(\arule_t)$.
Then, we propose the following augmented optimization objective:
\begin{align}\label{eq:RCA_objective_new}
    \max_\pi \mathbb{E}_\pi  \tilde{J}(\pi)\!:=\!\left[ \sum_{t=1}^h \!\gamma^{t-1}\! \tilde{r}_t \right] s.t. \text{ constraint~in} ~\eqref{eq:RCA_objective}.
\end{align}

We emphasize that LLMs cannot fully replace the ultimate human assessment, but they they provide a scalable alternative during the optimization process.

\section{Rule-based Reinforcement Learning (\rbrl)}
\label{sec:rbrl}

\rbrl  leverages the strengths of LLMs and RL to achieve both interpretability and robust sequential decision-making.

\textbf{Algorithm Overview}\quad
The \rbrl framework in ALgorithm \ref{alg:rbrl} involves four steps: (1) \textsc{Rule Set Generation} (line 3), the LLM processes the state-task $\bp_t$ to create candidate rules $\cR_t$ for action selection; (2) \textsc{Rule Selection} (line 4), an attention-based RL policy $\pi_{\theta}$ selects the best rule $\arule_t\in\cR$; (3) \textsc{Decision, Rule Reward and Explanation} (lines 5-8), the LLM generate an environment action $\aenv_t$ and based on the chosen rule $\arule_t$ and gives a human-readable explanation $\pmb{\ell}_t^{expl}$; (4) \textsc{Reinforcement Learning} (lines 10-12), collected samples update the policy $\pi_\theta$ and value networks with the SAC algorithm \cite{haarnoja2018soft} and the combined environment and rule reward $\tilde{r}_t$. Algorithm \ref{alg:rbrl} details the entire process. Further sections elaborate on these steps.

\begin{algorithm}[h]
\caption{\rbrl}
\begin{algorithmic}[1]
\REQUIRE 
Rule-selection policy $\pi_\theta$; Value network $\{Q^j_\phi\}_{j\in\{1,2\}}$; and Replay buffer $\cB$.

\STATE \textbf{Initialization:} Initial state $\bs_0$ and task-state prompt $\bp_0$.
\FOR{$t=0, \ldots, \texttt{max\_iters} - 1$}
\STATE Generate rule candidates $\cR_t$ using CoT from $\bp_t$ and $\athought_t$. {// Section \ref{sec:rulegen}}
\STATE Select rule $\arule_t$ using the RL policy $\pi_\theta$ from $\cR_t$ and $\bs_t$. {// Section \ref{sec:ruleselection}}
\STATE Generate the environment action $\aenv_t$ with the LLM from $\arule_t$, $\bp_t$, and previous thoughts.
\STATE Apply the action in the environment and obtain new state $\bs_{t+1}$ and environment reward $\renv_t$.
\STATE Generate post-hoc explanation with the LLM from $\aenv_t$, $\rrule_t$, $\bp_t$, and previous thoughts.
\STATE Generate rule reward $\rrule$ with the LLM as judge. {// Section \ref{sec:rulerew}}
\STATE Update the prompt $\bp_{t+1}$ from $\bs_{t+1}$, and the constraints $C$ and budget $B$.
\STATE Append transition to the replay buffer $\cB \leftarrow \cB \cup\{(\tilde{\bs}_t, \arule_t, \tilde{r}_t, \tilde{\bs}_{t+1})\}$ where $\tilde{\bs}_t=(\bs_t, \cR_t)$ and $\tilde{r}_t=r^{\text{rule}}_t + r^{\text{env}}_t$.

\IF {$t\mod k=0$}
\STATE Sample from the replay buffer and use Soft-Actor Critic RL to update the policy network $\pi_\theta(\arule_t | \tilde{\bs}_t)$ and value networks $\{Q^j_\phi(\tilde{\bs}_t, \arule_t)\}_{j\in\{1,2\}}$. {// Section \ref{sec:sac-rl}}
\ENDIF
\ENDFOR
\label{alg:rbrl}
\end{algorithmic}
\end{algorithm}

\begin{figure*}
    \centering
    \includegraphics[width=\linewidth]{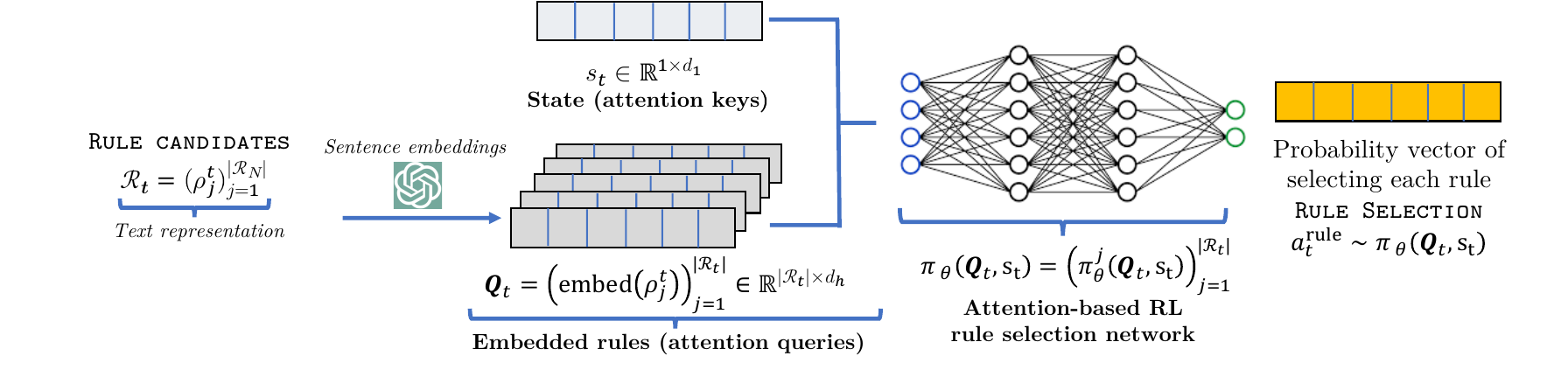}
    \caption{Overview of the \textsc{Rule Selection} step. The current state is encoded as a key vector, while candidate rules are encoded as Queries using a text embedding API (e.b., BERT sentence embedding). An attention-based policy network $\pi_\theta$ trained with SAC computes a probability distribution over the candidate rules, enabling the selection of the most suitable rule for decision-making and explanation.}
    \label{fig:rule_selection}
\end{figure*}

\subsection{Rule Set Generation}
\label{sec:rulegen}

The rule generation process seeks to to create interpretable and actionable guidelines for decision-making.
Under this framework, a set of candidate rules $\cR_t$ is generated according to $\cR_t\sim \pi_{\llm}(\cR_t|\bp_t, \athought_t)$. To enhance interpretability, each rule is accompanied by a rationale explaining the reasoning behind the decision.  The LLM is instructed to generate rules as a JSON format, which is common for integration of LLMs with downstream applications \cite{together2024function}. Examples of generated rules are given in Figure \ref{fig:rule_prompt}. See Figure \ref{fig:combined_prompt} in the Appendix for the prompt templates used to generate the rules.

 A higher number of rules is always preferable, however, it also increases the generating costs and slows down each iteration of the method. For our experiments, we found that $|\cR_t|=5$ provided achieved a reasonable trade-off.

\subsection{Rule Selection}\label{sec:ruleselection}

In this step, rules are converted from text to vector form, and a trainable attention-based policy network $\pi_{\theta}$ provides the probability distribution for sampling a rule. Figure \ref{fig:rule_selection} illustrates the process, with a detailed procedure in Algorithm \ref{alg:rule_search} of the Appendix. Below are the major components of the architecture of $\pi_\theta$. We propose to base the architecture on cross-attention layers \cite{bahdanau2014neural, vaswani2017attention}, with the state acting as the keys and values, and the rules as the queries. This allows to learn from the embedding representations of rules, and efficiently handle dynamically changing number of rules if needed.

\textit{State Representation.} The numeric state is projected by a linear layer: $\mathbf{k}_t =\texttt{Linear}(\bs_t) \in \mathbb{R}^{d_h}$, with $d_h$ being to denote the architecture hidden dimension. 

\textit{Rule Candidate Embedding.}
Each rule in the {list of rule candidates} $\mathcal{R}_t = \{ \pmb{\rho}^t_1, \pmb{\rho}^t_2, \dots, \pmb{\rho}^t_q \}$ is embedded into a numeric representation using a {Sentence Embedding} language model (e.g., SentenceBERT \cite{reimers2019sentencebert}) and further processed by a projection layer similar to the state representation. This results in a \textit{query} matrix
$\mathbf{Q}_t \in \mathbb{R}^{q \times d_h}$.

\textit{Attention-based Policy Network.}  The vector $ \mathbf{k}_t$, serving as keys, engages with the rule embeddings $\mathbf{Q}_t$, acting as queries, via a cross-attention mechanism to derive a hidden state representation $\bh_t^{(1)}=\texttt{Attention}(\mathbf{Q}_t, \bk_t^\top, \bk_t^\top)\in\mathbb{R}^{q \times d_h}$, computed as
$$
\begin{aligned}
\texttt{Attention}(\mathbf{Q}_t, \bk_t^\top, \bk_t^\top) = \text{softmax} \left(\frac{\mathbf{Q}_t \bk_t}{\sqrt{d_h}} \right) \bk_t^\top,
\end{aligned}
$$
which facilitates the rule candidate vector embeddings in attending to the environment state. Subsequently, we sequentially apply self-attention layers to the hidden representation $\bh^{(k+1)}=\texttt{Attention}(\bh_t^{(k)}, \bh_t^{(k)}, \bh_t^{(k)})$, enabling the rule embeddings to attend to one another to rank an optimal candidate. Ultimately, following $K-1$ self-attention layers, a final linear layer converts the rule representations into logit vectors $\pmb{\alpha}^t_{\theta}=\texttt{Linear}(\bh_t^{(k)})\in\mathbb{R}^q$ used for the computation of the probability of selecting each rule.

For the implementation, the attention layer is realized using the multi-headed attention module from \citet{vaswani2017attention}. We incorporate a dropout layer, fixed at 0.05 for the experiments, along with SiLU activation and layer normalization, which are excluded from the notation for brevity.

\textit{Rule Selection.}
The policy distribution over the rules:
$$
\pi_{\theta, i}(\mathbf{Q}_t,\mathbf{k}_t) = \frac{\exp(\alpha^t_{\theta, i}(\mathbf{Q}_t,\mathbf{k}_t))}{\sum_{j=1}^q \exp(\alpha^t_{\theta, j}(\mathbf{Q}_t,\mathbf{k}_t))}, \quad i=1,\ldots, q.
$$
A rule is selected sampled from the distribution:
$\arule_t \sim \texttt{Categorical}(\cR; (\pi_{\theta,i}(\mathbf{Q}_t, \mathbf{k}_t))_{i=1}^q)$.

\subsection{Decision, Rule Reward, and Explanation} \label{sec:rulerew}
Upon selection of rule $\arule_t$, the LLM determines the action to be applied within the environment $\aenv_t \sim \pi_\llm(\aenv_t | \arule_t, \athought_t, \bp_t)$, ensuring concordance with the chosen strategy. Subsequently, the LLM formulates a post-hoc explanation $\aexpl_t \sim \pi_\llm(\aexpl_t | \aenv, \arule, \athought, \bp_t)$ contingent upon the rule. Figure \ref{fig:combined_prompt} in the Appendix illustrates the prompt template employed to generate both the action and explanation. 

This procedure concurrently produces the rule reward $\Rrule_\llm(r^{\text{rule}}_t)$, used for reinforcement learning (RL) in the next step. This rewards is derived from using the LLM as a judge to answer the following three questions:

$\texttt{ER}_1$. Without providing $\aenv_t$, is $\arule_t$ sufficient to predict/infer the optimal action?

$\texttt{ER}_2$. Does $\arule_t$ contain enough details about the applicability of the rule to the current state?

$\texttt{ER}_3$. Given $\aenv_t$, is $\arule_t$ compatible with this selection?

Each question scores as $0$ if negative or $1$ if positive. The rule reward is calculated as $r_t^\text{rule}=\Rrule_\llm(\arule_t)=(1/3)\sum_i \texttt{ER}_i$. Refer to Figure \ref{fig:combined_prompt} in the Appendix for the full prompt.

\subsection{RL with SAC}\label{sec:sac-rl}

\emph{Augmented state space}\quad Traditional RL frameworks fail here due to intermediate steps: generating the rule set \(\cR_t\), mapping rules \(\arule_t\) to actions \(\aenv_t\) in an LLM-driven environment. \rbrl addresses this issue by creating an augmented state \(\tilde{\bs}_t := (\bs_t, \cR_t)\) with transition dynamics \(P(\tilde{\bs}_{t+1} | \tilde{\bs}_t, \arule_t)\), integrating rules into the state space for reasoning over both the environment's dynamics and decision rules \(\arule_t\). This proposition explains the transition computation.
\begin{theorem}
The state transition of the \rbrl MDP  can be calculated as 
\begin{align}
&P(\tilde{\bs}_{t+1}|\tilde{\bs}_t, \arule_t)=P( \cR_{t+1}|\bs_{t+1})\times\nonumber\allowdisplaybreaks\\ &\qquad\int_\ba P(\bs_{t+1}|\aenv,\bs_t)
    \cdot P(\aenv| \arule_t, \bs_t)d\aenv,
\end{align}
where $P( \cR_{t+1}|\bs_{t+1})= \pi_\llm(\cR_{t+1} | \bp_t, \pmb{\tau}_t)$ is the probability of the LLM generating rule set $\cR_{t+1}$ provided the state $\bs_{t+1}$, $P(\bs_{t+1}|\aenv,\bs_t)$ is the original environment dynamics, and $P(\aenv| \arule_t, \bs_t)=\pi_\llm(\aenv | \bp_t, \arule_t)$ is the probability of the LLM selecting the environment action $\aenv$.
\label{prop:transition}
\end{theorem}

\emph{SAC step}\quad The attention-based policy network in Section \ref{sec:ruleselection} is optimized using the SAC algorithm, which balances reward maximization with exploration by incorporating an entropy term in the objective function. The SAC update process is outlined as follows.
First, we define an auxiliary target value:
\begin{align}
    y&=\Big( \tilde{r}_t + \gamma \mathbb{E}_{\arule_{t+1} \sim \pi_\theta} \Big[ \min_{j=1,2} \bar{Q}_{\bar{\phi}_j}(\tilde{\bs}_{t+1}, \arule_{t+1})\nonumber\\
    &\qquad\qquad\qquad- \beta \log \pi_\theta(\arule_{t+1}|\bQ_{t+1}, \bk_{t+1}) \Big] \Big),
\end{align}
The Q-networks are updated by minimizing the loss function: 
\begin{align}
    L_Q(\phi_i) &= \mathbb{E}_{\mathcal{D}} \Big[ \Big( Q_{\phi_i}(\tilde{\bs}_t, \arule_t) -y \Big)^2 \Big], \forall i=1,2.
\end{align}
The policy network is updated by minimizing the KL divergence between the policy and the Boltzmann distribution induced by \( Q_\phi \), which is expressed as
    \begin{align}
    \hspace{-0.2cm}
        L_\pi(\theta)\!\!=\!\! \mathbb{E}_{\cD} \!\Big[\! \beta \!\log \!\pi_\theta(\arule_t | \bQ_t, \!\newblock{}_t)\!\! -\! \!\min_{i=1,2}\!Q_{\phi_i}\!(\tilde{\bs}_t, \arule_t) \!\Big],
    \end{align}
where $\beta$ is a temperature parameter.  The full SAC update procedure is detailed in Algorithm \ref{algo:SAC} in Appendix \ref{Sec:algo_appendix}.

\section{Experiments \& Human Survey}\label{sec:exp}

\begin{figure*}[!th]
    \centering
    \begin{subfigure}[b]{0.33\linewidth}
        \centering
        \includegraphics[width=\linewidth]{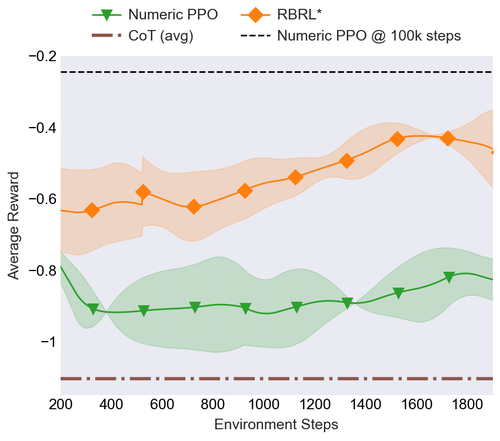}
        \vskip -4pt
        \caption{\texttt{Uganda} environment}
        \label{fig:uganda-exp}
    \end{subfigure}
    \begin{subfigure}[b]{0.33\linewidth}
        \centering
        \includegraphics[width=\linewidth]{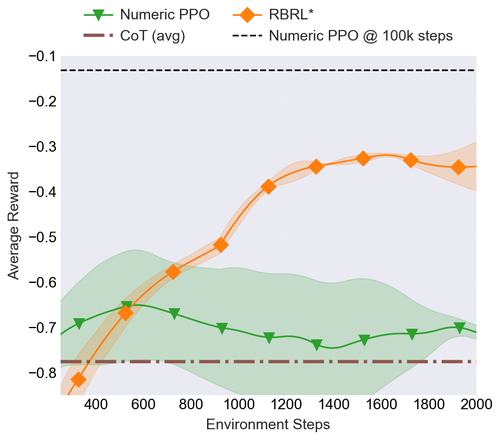}
        \vskip -4pt
        \caption{\texttt{MimicIII} environment}
        \label{fig:mimic-iii-exp}
    \end{subfigure}
    \begin{subfigure}[b]{0.33\linewidth}
        \centering
        \includegraphics[width=\linewidth]{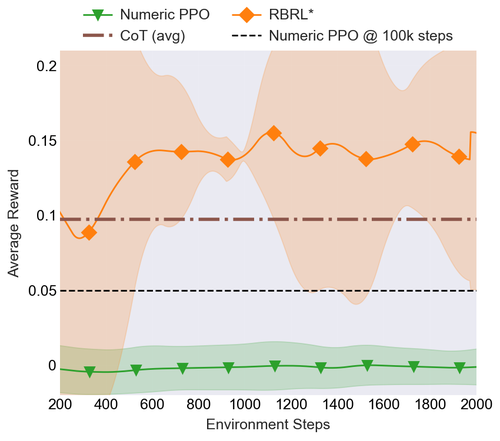}
        \vskip -4pt
        \caption{\texttt{HeatAlerts} environment}
        \label{fig:mimic-iii-exp}
    \end{subfigure}
    \vskip -4pt
    \caption{Results from Q1. Main comparison of RBRL on three resource allocation problems. The plots show the mean and standard error across three seeds, using exponentially weighted moving averages with a half-life of 100.}
    \label{fig:results-reward}
\vskip -6pt
\end{figure*}

\begin{figure*}[!th]
    \centering
    \begin{subfigure}[t]{0.34\linewidth}
        \centering
        \includegraphics[width=1\columnwidth]{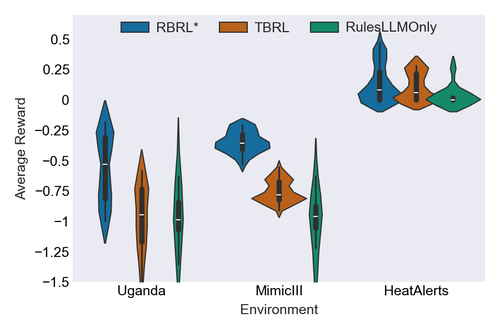}
        \vskip -4pt
        \subcaption{Results from Q2: ablations}
        \label{fig:tbrl}
    \end{subfigure}
    \begin{subfigure}[t]{0.3\linewidth}
        \centering
\includegraphics[width=1\columnwidth]{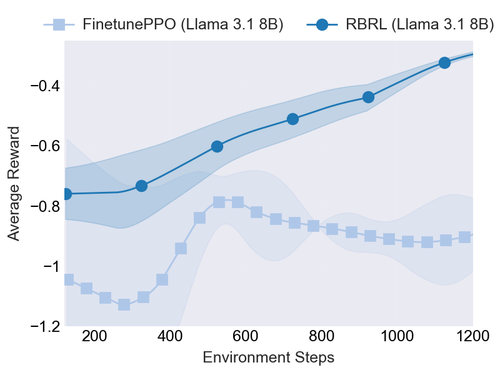}
    \vskip -4pt
\subcaption{Results from Q3: LLM finetuning}
 \label{fig:finetune}
    \end{subfigure}
    \begin{subfigure}[t]{0.34\linewidth}
        \centering
        \includegraphics[width=1\columnwidth]{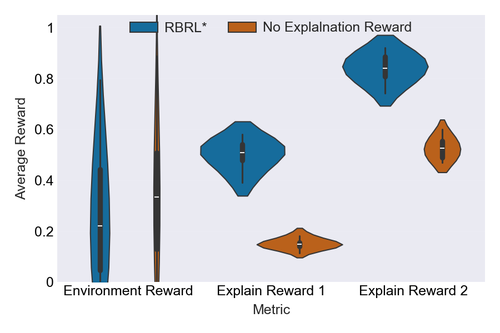}
                \vskip -4pt
\subcaption{Results from Q4: no rule rewards}
        \label{fig:no-rulerew}
    \end{subfigure}

    \vskip -4pt
    \caption{Additional experiments and ablations. (a) Comparison of \rbrl with thoughts-based RL (TBRL) and the baseline rule-based LLM without RL training; (b) comparison against LLM finetuning with PPO at the token level from the environment reward with CoT generation for the \texttt{Mimic environment}; (c) shows the effect of removing the rule reward in the \texttt{HeatAlerts} environments. For (a) and (c), we show distribution of rewards in the last 20\% training steps.}
    \label{fig:additional-restuls}
\end{figure*}

In this section, we evaluate \rbrl and empirically show that it can achieve a joint improvement in both reward and explainability over comparable baselines
. We briefly summarize these environments here, with additional details in Appendix \ref{appdx:env-details}.

\paragraph{Domains} We evaluate \rbrl in three environments pertaining to the following two distinct domains:

$\bullet$ \texttt{WearableDeviceAssignment}: We use two environments, \texttt{Uganda} and \texttt{MimicIII}, from the vital sign monitoring domain introduced by \citet{boehmer2024optimizing}, modeling the allocation of limited wireless devices among postpartum mothers as an restless multi-armed bandit problem (RMAB). Each mother (``arm") is represented by a state of vital signs (e.g., heart rate, respiratory rate, SPO2) and their variability. Action vector elements are binary: assigning a device to a mother or not. The aim is to prioritize high-risk patients since there are only a limited number of $B$ devices. Rewards penalize abnormal vital signs, with monitoring enabling health-improving interventions.
 \citet{boehmer2024optimizing} suggests to recast the RMAB problem as an MDP, deciding from which patient to remove a device when a new patient arrives.  A penalty is incurred when removing the device from an active patient if there are free devices.

$\bullet$ \texttt{HeatAlerts}: We use the \texttt{weather2alert} environment from \citet{considine2023optimizing}
, which formulates issuing heat alerts as a constrained MDP. The action is binary, subject to a constraint on the total number of alerts that can be issued. The state includes the heat index, the remaining budget, and a history of previous alerts. Actions $\ba_t$ are binary and represent whether to issue an alert. Rewards are reductions in hospitalizations from the alert. A penalty is incurred when selecting to issue an alert when on budget. 

\subsection{Environment Reward Optimization}

We discuss the main results and refer to Appendix \ref{appdx:setup} for the detailed experiment setup. Unless otherwise specified, we use \texttt{gpt-4o-mini} \cite{openai2024gpt4omini} as LLM due to its reasonable cost and high performance. All baselines were trained on 3 seeds for 2000 environment steps each. 

\emph{Q1. Did \rbrl optimize the reward function?} \quad \rbrl is compared to Chain-of-thought (CoT) \cite{wei2022chain} for language reasoning and Proximal Policy Optimization (PPO) \cite{schulman2017proximal} for numeric states. Figure \ref{fig:results-reward} indicates \rbrl outperforms CoT, showing RL-optimized rule selection improves decision-making. 
\rev{
    The results show that \rbrl exceeds PPO in all environments with equivalent environment steps, suggesting a higher performance when learning in an online setting. For completeness, we also compare against PPO trained at 100k steps, 50 times more. For the \texttt{HeatAlerts} environment, \rbrl exceeds such asymptotic performance, consistently with the findings of \citet{considine2023optimizing} noting that PPO and other standard RL algorithms struggle to handle the constraints and exploration in this domain. For the \texttt{Uganda} and \texttt{MimicIII} environments, we observe the trend of \rbrl getting closer to 100k PPO, but not reaching the performance. As highlighted in Section \ref{sec:related}, various works have remarked that the trade-off between interpretability and performance can be justified to increase system trust, robustness, and adaptability in high-stakes environments.
}

\emph{Q2. Did structured rules help optimization?} \quad We conduct two ablation studies on structured rules. First, we benchmark the use of structured rules without RL, called baseline \texttt{RulesLLMOnly}, which is shown in Equation (2)-(5). Next, we compare \rbrl with a variant optimizing unstructured thoughts, termed thoughts-based RL (\texttt{TBRL}). The implementation mimics \rbrl, utilizing a candidate pool $\cR$ with the CoT prompt. Results in Figure \ref{fig:tbrl} show that comparing \rbrl with \texttt{RulesLLMOnly} highlights RL training gains, suggesting rule generation alone does not explain \rbrl's performance. Additionally, significant improvements over \texttt{TBRL} suggest optimizing structured rules is more effective than optimizing free reasoning traces.

\emph{Q3. How does \rbrl compare to token-level LLM finetuning with RL?} \quad  We implement LLM finetuning from the environment reward to a Llama 3.1 8B model \cite{meta2024llama3.1}, termed \texttt{FinetunePPO}. We train a value head on the last hidden states and use KL divergence from the reference model as a regularization reward \cite{ziegler2019fine}. A simple CoT generation is used, followed by a call for action question, optimizing the full CoT and action trace. We train 3 seeds for 18 hours on an Nvidia A100 40G GPU, achieving 1200 steps per seed. For compatibility, we also train \rbrl on the same Llama 3.1 8B model. Figure \ref{fig:finetune} compares results, showing a positive but relatively flat trend for finetuning compared to \rbrl, suggesting \rbrl is better online. Additionally, \rbrl runs on a regular laptop, unlike \texttt{FinetunePPO} that needs specialized hardware, and training time for \texttt{FinetunePPO} is about 4x longer on equivalent steps. Due to computation constraints, we only show this experiment for the \texttt{MimicII} domain, which had the less noise in the previous experiments.

\subsection{Human Survey and Explainability}

\begin{figure}[tbp]
\centering
\includegraphics[width=0.85\columnwidth]{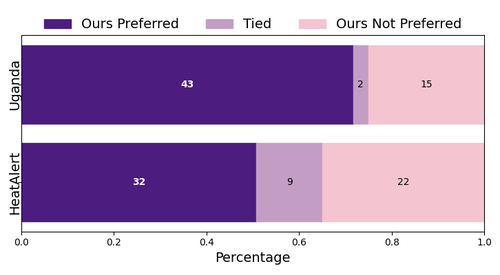}
\vskip-10pt
\caption{Results from the human survey.}
\label{fig:survey}
\vskip -8pt
\end{figure}

\emph{Q4. Did \rbrl increase the explainability of post-hoc explanations?}  \quad 
A survey with 40 participants was conducted to assess explanation quality, detailed in Appendix \ref{appdx:survey}. Each prompt included the task, state, and action space as originally given to the LLM, followed by actions and explanations from the CoT agent and the \rbrl agent, without disclosing agent types. Participants were asked to choose preference for explanation A, B, or none. Prompts were split between \textit{Wearable Device Assignment} and \textit{Heat Alerts} domains. Figure \ref{fig:survey} shows results, favoring \rbrl's explanations in both domains, with a detailed breakdown in \ref{appdx:survey}. An additional experiment with an LLM judge \cite{gu2024survey} using a large \texttt{gpt-4o} model \cite{openai2024gpt4o} showed strong agreement with humans, preferring \rbrl's explanations in all but one case.

\emph{Q5. What was the effect of the rule reward?}  During training of \rbrl, rules received rewards based on two prompts. We examine an ablation where this reward was omitted. Figure \ref{fig:no-rulerew} illustrates results for the \texttt{HeatAlerts} environment, noted for high variance and a challenging reward function. We extended training to 5k environment steps to better understand these dynamics. When \rbrl is trained without the rule reward, the environment reward remains steady (with a slight increase), but explainability scores drop significantly. Refer to Section \ref{sec:rulerew} for the definition of the rule reward metrics. A decline in metric 1 indicates that rules are less predictive of the optimal actions. A decline in metric 2 suggests rules lack detailed applicability to the current problem state, indicating more generic rather than specialized rule selection. Metric 3 (not shown) was always 1 in all steps, indicating the limitations of directly evaluating post hoc explanations.  
Although judged by the LLM, these results are encouraging, as our previous experiment showed alignment between the LLM and human assessments.

\section{Conclusion}
\label{sec:conclusion}

The \rbrl framework takes an important step towards addressing the critical challenge of balancing decision efficacy and interpretability in constrained allocation tasks. By synergizing the generative capabilities of LLMs with the optimization power of RL, \rbrl introduces a novel paradigm where structured rules, generated dynamically by LLMs, guide RL policies to make transparent and actionable decisions. Evaluations across healthcare and public health domains demonstrate \rbrl’s competitive performance against traditional RL baselines while significantly enhancing the quality of human-understandable explanations. Human and LLM-judged surveys further validate that RBRL’s explanations improve trust and clarity, contributing towards addressing the longstanding barrier to deploying AI in high-stakes scenarios.

\section*{Impact and Ethics Statement}
This work advances the development of transparent AI systems for high-stakes decision-making in domains like healthcare and public policy. By enabling RL agents to generate human-readable rules and explanations, \rbrl improves trust and accountability, critical for ethical deployment in settings where lives and resources are at stake.

While the framework prioritizes alignment with human reasoning, potential risks include over-reliance on imperfect LLM-generated rules or explanations that may inadvertently obscure biases in training data. Mitigation requires rigorous validation of rules by domain experts and ongoing monitoring of LLM outputs. Additionally, \rbrl’s reliance on LLMs raises computational and accessibility challenges in resource-constrained environments. By addressing these considerations, this research contributes to safer, more equitable AI systems that empower—rather than replace—human decision-makers.

Notice that the Uganda and Heat Alerts datasets used in this study is derived from a simulator provided by \citet{boehmer2024optimizing} and \citet{considine2023optimizing}. These simulators do not include any feature information or identifying details of real patients. Thus, the data generated by the simulator cannot be traced to or represent actual individuals, ensuring privacy and ethical compliance. We emphasize that this is purely a simulated study; and recognize that for any next steps towards real world use, there is a need to conduct rigorous simulation studies on a large scale with real patient data, with detailed assessments of potential biases, verification of policy convergence and its robustness to distribution shifts in patient populations, and making necessary adjustments. Beyond that, there will be a need to obtain ethics and regulatory approval to test the policy in a real-world setting for future comprehensive field-testing, addressing issues of participant consent and privacy; and ultimately there would need to be sufficient human oversight for any future deployment.

\rev{
\section*{Acknowledgments}  
This work is supported by the Harvard Data Science Initative, Wadhwani AI, and the National Institues of Health (R01ES34021,
R01ES037156.)
}

\bibliography{references}
\bibliographystyle{icml2025}

\newpage
\appendix
\onecolumn

\section{Algorithmic and Mathematical Details}

\subsection{Algorithms}\label{Sec:algo_appendix}

In this subsection, we present the detailed pseudocodes for \texttt{Rule\_Search} in Algorithm \ref{alg:rule_search} and the SAC for attention-based policy network in Algorithm \ref{algo:SAC}.

\begin{algorithm}[H]
\caption{\texttt{Rule\_Search}: Rule Selection via Attention-Based Policies}
\begin{algorithmic}[1]
\REQUIRE Numeric state representation $\bs_t \in \mathbb{R}^{d_s}$  and rule set $\cR_t=\{\pmb{\rho}^t_{i}\}_{i=1}^q$. Hidden dimension $d_h$.

\STATE Embed each rule \( \pmb{\rho}^t_{i} \in \cR_t \) into a numeric vector using sentence embeddings $\bq^t_i=\texttt{embed}(\pmb{\rho}^t_i)\in\mathbb{R}^{d_h}$ (e.g., Sentence-BERT) and stack to form a query matrix \( \mathbf{Q}_t \in \mathbb{R}^{d_h \times q} \). {// \textit{Rule Candidate Embedding}}

\STATE The state $\bs_t$ is projected by a linear layer with SiLU activation: $\mathbf{k}_t =\texttt{SiLU}(\texttt{Linear}(\bs_t)) \in \mathbb{R}^{d_h}$, with $d_h$ being to denote the architecture hidden dimension. { // \textit{State Representation}}

\STATE Use cross-attention to obtain $\bh=\mathrm{CrossAttention}(\mathbf{Q}_t, \mathbf{k}_t, \mathbf{k}_t)\in \mathbb{R}^{q \times d_h}$.

\STATE (Optional) Further apply a self-attention network $\bh \leftarrow \mathrm{SelfAttention}(\bh)$.

\STATE Apply linear layer to obtain logits vector $\pmb{\alpha}_{\theta}^t=\mathrm{Linear}(\bh) \in \mathbb{R}^{q \times 1}$.

\STATE Calculate the policy distribution:
$\pi_{\theta, i}(\mathbf{Q}_t,\mathbf{k}_t) = \frac{\exp(\alpha^t_{\theta, i}(\mathbf{Q}_t,\mathbf{k}_t))}{\sum_{j=1}^q \exp(\alpha^t_{\theta, j}(\mathbf{Q}_t,\mathbf{k}_t))}$, where $\alpha^t_{\theta, i}$ is $i$-th element in $\pmb{\alpha}_{\theta}^t$.

\STATE Sample rule $\arule_t \sim \texttt{Categorical}(\cR; (\pi_{\theta,i}(\mathbf{Q}_t, \mathbf{k}_t))_{i=1}^q)$. {// \textit{Rule Selection with Attention}}
\STATE \textbf{Return} $\arule_t$.
\label{alg:rule_search}
\end{algorithmic}
\end{algorithm}

Algorithm \ref{alg:rule_search} outlines the process of rule selection using attention-based policies.   First, each rule candidate $\pmb{\rho}_i^t$ is embedded into a numeric vector $\bq_i^t$ using a sentence embedding technique (e.g., Sentence-BERT), forming a query matrix $\bQ_t$. The state $\bs_t$ is also converted into a numeric vector $\bk_t$. Cross-attention is applied between $\bQ_t$ and $\bk_t$ to generate an attention representation  $\bh$, which may optionally be refined using a self-attention mechanism. A linear layer processes to $\bh$ produce score vector $\pmb{\alpha}_{\theta}^t$. These scores define the policy distribution $\pi_\theta$, from which a rule $\arule_t$
is sampled. This attention-based approach ensures efficient selection of rules by leveraging contextual relationships between the state and rule candidates.

\begin{algorithm}
\caption{SAC for Attention-based Policy Network}
\label{algo:SAC}
\begin{algorithmic}[1]
\STATE Initialize Q networks $Q_{\phi_1}$, $Q_{\phi_2}$ and policy network $\pi_\theta$ with random parameters $\phi_1$, $\phi_2$, $\theta$.
\STATE Initialize target Q networks $\bar{Q}_{\bar{\phi}_1}$, $\bar{Q}_{\bar{\phi}_2}$ with weights $\bar{\phi}_1 \leftarrow \phi_1$, $\bar{\phi}_2 \leftarrow \phi_2$.
\STATE Initialize temperature parameter $\beta$ and target entropy $\mathcal{H}_{\text{target}}$;
 Initialize replay buffer $\mathcal{D}$.

\FOR{episode $= 1, \dots, M$}
    \STATE Initialize environment and observe initial state $\tilde{\mathbf{s}}_1$.
    \FOR{step $t = 1, \dots, T$}
        \STATE Sample action $\arule_t \sim \pi_\theta(\cdot|\tilde{\bs}_t)$.
        \STATE Execute action $\arule_t$, observe reward $\tilde{r}_t$ and next state $\tilde{\bs}_{t+1}$.
        \STATE Store transition $(\tilde{\bs}_t, \arule_t, \tilde{r}_t, \tilde{\bs}_{t+1})$ in replay buffer $\mathcal{D}$.
        
        \IF{enough samples in $\mathcal{D}$}
            \STATE Sample a mini-batch of transitions $(\tilde{\bs}_t, \arule_t, \tilde{r}_t, \tilde{\bs}_{t+1})$ from $\mathcal{D}$.
            
            \STATE Compute target Q values: $y_t = \tilde{r}_t + \gamma\mathbb{E}_{\arule_{t+1}\sim\pi_{\theta}(\cdot|\tilde{\bs}_{t+1})}\left[  \min_{j=1,2} \bar{Q}_{\bar{\phi}_j}(\tilde{\bs}_{t+1}, \arule_{t+1}) - \alpha \log \pi_\theta(\arule_{t+1}|\tilde{\bs}_{t+1}) \right]$.
            
            \STATE Update Q networks by minimizing:
            \STATE $L_Q(\phi_i) = \mathbb{E}_{(\tilde{\bs}_t, \arule_t, \tilde{r}_t, \tilde{\bs}_{t+1})} \left[ \left( Q_{\phi_i}(\tilde{\bs}_t, \arule_t) - y_t \right)^2 \right]$ for $i = 1, 2$.
            
            \STATE Update policy network by minimizing:
            \STATE $L_\pi(\theta) = \mathbb{E}_{\tilde{\bs}_t, \arule_t \sim \pi_\theta} \left[ \beta \log \pi_\theta(\arule_t|\tilde{\bs}_t) - \min_{j=1,2} Q_{\phi_j}(\tilde{\bs}_t, \arule_t) \right]$.
            
            \STATE Update temperature parameter by minimizing:
            \STATE $L_\beta(\beta) = \mathbb{E}_{\tilde{\bs}_t, \arule_t \sim \pi_\theta} \left[ -\beta \left( \log \pi_\theta(\arule_t|\tilde{\bs}_t) + \mathcal{H}_{\text{target}} \right) \right]$.
            
            \STATE Update target Q networks:
            \STATE $\bar{\phi}_i \leftarrow \tau \phi_i + (1 - \tau) \bar{\phi}_i$ for $i = 1, 2$.
        \ENDIF
    \ENDFOR
\ENDFOR
\end{algorithmic}
\end{algorithm}

Algorithm~\ref{algo:SAC} presents the SAC algorithm tailored for training an attention-based policy network in selecting the desired rule. This method combines entropy-regularized policy optimization with a structured approach to handle rule-selection effectively.
The algorithm begins with the initialization of key components: Q networks \( Q_{\phi_1}, Q_{\phi_2} \), target Q networks \( \bar{Q}_{\phi_1}, \bar{Q}_{\phi_2} \), and a policy network \( \pi_\theta \). Random parameters are assigned to these networks, and the target Q networks are synchronized with the initial Q networks. A temperature parameter \( \alpha \) is initialized to regulate the entropy $\cH_{target}$ in the policy objective, ensuring a balance between exploration and exploitation. A replay buffer \( \mathcal{D} \) is set up to store transition data. Notice the entropy is defined as
\begin{align}
    \mathcal{H}_{target} = - \sum_{i=1}^q \pi_\theta(\arule_t | \bk_t, \mathbf{Q}_t) \log \pi_\theta(\arule_t | \mathbf{k}_t, \mathbf{Q}_t).
\end{align}

During training, each episode starts with the initialization of the environment, and the agent observes the initial state \( \tilde{\bs}_1 \). At every time step, the policy network generates an action \( \arule_t \) based on the current state. This action is executed in the environment, resulting in a reward \( \tilde{r}_t \) and a state transition to \( \tilde{\bs}_{t+1} \). These transitions are stored in the replay buffer for optimization.
When sufficient transitions are available in the buffer, the algorithm samples a mini-batch of transitions and computes the target Q values. The target Q values incorporate entropy regularization and are computed using the minimum of the target Q networks to ensure stability. The Q networks are updated by minimizing the mean squared error between the predicted Q values and the computed targets.
The policy network is optimized by minimizing a loss function that combines the policy entropy with the expected Q value, ensuring a stochastic and exploratory policy. The temperature parameter $\beta$ is updated to maintain the desired balance between exploration and exploitation. Finally, the target Q networks are softly updated to stabilize training. This iterative process continues across episodes and time steps, progressively refining the policy network to achieve optimal rule selection.

\subsection{Proof of Theorem \ref{prop:transition}}
In this section, we provide the detailed proofs for Theorem \ref{prop:transition}.
We start with the following equation
\begin{align}
P( \cR_{t+1}|\bs_{t+1})&\cdot \int_a P(\bs_{t+1}|\aenv,\bs_t)
    \cdot P(\aenv|\arule_t, \bs_t)d\aenv\nonumber\allowdisplaybreaks\\
   &= \underset{(a)~ {\color{blue} P(\bs_{t+1}|\aenv,\bs_t, \cR_t, \arule_t)=P(\bs_{t+1}|\aenv,\bs_t)}}{\underbrace{P( \cR_{t+1}|\bs_{t+1})\cdot\int_a P(\bs_{t+1}|\aenv,\bs_t, \cR_t, \arule_t)\cdot P(\aenv|\cR_{t}, \arule_t, \bs_t)d\aenv}}\nonumber\allowdisplaybreaks\\
   &=\underset{(b)~{\color{blue}P(\aenv|\bs_t,\cR_{t}, \arule_t)=P(\aenv|\cR_{t}, \arule_t)}}{\underbrace{ P( \cR_{t+1}|\bs_{t+1})\cdot\int_a P(\bs_{t+1}|\aenv,\bs_t, \cR_t, \arule_t)\cdot P(\aenv|\bs_t,\cR_{t}, \arule_t)d\aenv}}\nonumber\allowdisplaybreaks\\
   &= \underset{P(A|B,C)\cdot P(B|C)=P(A, B|C)}{\underbrace{P( \cR_{t+1}|\bs_{t+1})\cdot\int_a P(\bs_{t+1},\aenv|\bs_t, \cR_t, \arule_t)d\aenv}}\nonumber\allowdisplaybreaks\\
   &=P( \cR_{t+1}|\bs_{t+1})\cdot P(\bs_{t+1}|\bs_t, \cR_t, \arule_t)\nonumber\allowdisplaybreaks\\
   &=\underset{(c)~{\color{blue}P( \cR_{t+1}|\bs_{t+1}, \bs_t, \cR_t, \arule_t)=P( \cR_{t+1}|\bs_{t+1})}}{\underbrace{P( \cR_{t+1}|\bs_{t+1}, \bs_t, \cR_t, \arule_t)\cdot P(\bs_{t+1}|\bs_t, \cR_t, \arule_t)}}\nonumber\allowdisplaybreaks\\
    &=\underset{P(A|B,C)\cdot P(B|C)=P(A, B|C)}{\underbrace{P(\bs_{t+1}, \cR_{t+1}|\bs_t, \cR_t, \arule_t)}}\nonumber\allowdisplaybreaks\\
    &=\underset{\tilde{s}_t:=(\bs_t,\cR_t)}{\underbrace{P(\tilde{\bs}_{t+1}|\tilde{\bs}_t, \arule_t)}},
\end{align}
where $(a)$ follows from the fact that the transition to $\bs_{t+1}$ is fully determined by current state $\bs_t$ and current action to the environment $\aenv_t$, i.e., independent on rule set $\cR_t$ and selected rule $\arule_t$; $(b)$ holds since $\aenv_t$ is determined only by the selected rule $\arule_t$ and the state $\bs_t$; $(c)$ is due to our designed rule generation procedure where $\cR_{t+1}$ is generated by the LLM from the the latest state $\bs_{t+1}$.  
This completes the proof.

\section{Experiment Setup}
\label{appdx:setup}

\subsection{Environments Details}
\label{appdx:env-details}

\subsubsection{Wearable Device Assignment Domain}\label{appdx:health_care}

The simulator for the Uganda domain is adapted from \citep{boehmer2024optimizing} with minor modifications to simplify the problem.  This section provides an overview of the environment, with additional details available in the original paper.  In this environment, they want to allocate vital sign monitoring devices to mothers arriving in a maternal unit in order to better monitor mothers' health condition. Each mother's state is modeled by her vital signs (heart rate, respiratory rate, and blood oxygen saturation) along with each vital sign's variability. The mother's vital sign transition is governed by a multivariate Gaussian distribution defined over her vital signs at the current timestep and next timestep, learned from de-identified vital sign data collected from patients at Mbarara Regional Referral Hospital. MIMIC-III \cite{johnson2016mimic} is another de-identified clinical vital sign dataset that includes the same set of vital signs as the Uganda domain. The key difference is they have different data sources, as MIMIC-III's data comes from Beth Israel Deaconess Medical Center in Boston.

Wearing a monitoring device does not directly alter a mother's vital sign trajectory but has an indirect positive effect by triggering alerts when vital signs deviate from the normal range. These alerts often lead to medical interventions that improve the mother's condition.  If no monitoring device is assigned (passive action), the mother's next state is sampled from the multivariate Gaussian distribution conditioned on the current state. If a monitoring device is assigned and the vital signs remain within the normal range, the vital signs evolve as under the passive action. However, if any vital sign deviates from the normal range,  there is a $30\%$ chance the vital signs evolve as under the passive action, based on empirical evidence suggesting that clinicians fail to respond in such cases 30\% of the time \citep{boatin2021wireless}. Otherwise, vital signs are probabilistically adjusted towards the normal range before sampling the next state, modeling the positive impact of medical intervention.

The algorithm's goal is to optimize monitoring device allocation to maximize the aggregate reward across all mothers. We simplify the problem by requiring exactly one new mother to join the maternal unit at each timestep, starting with a single mother in the unit. The system has a budget of five monitoring devices. A device must be allocated to the new mother, and if all devices are already in use, one must be removed from a current user. Once removed, a device cannot be reassigned to the same mother. Each mother remains in the maternal unit for 10 timesteps, after which her vital sign trajectory no longer contributes to the reward. Once a device is taken from a mother, we directly sample her entire vital sign trajectory under passive action for the remaining timesteps she stays in the maternal unit and compute all her future rewards. We can directly compute future rewards because the mother will not receive the device again, so she will only undergo passive action in the remaining time. This observation enables us to maintain a smaller observation space, as we only need to keep track of the states of the mothers who own the device.

In this domain, the constraints can be written as  $\|\ba_t\in\mathcal{R}^{d_2}\|_1\leq B, \forall t$ , which $d_2$ represents the number of patients in the system at each time slot, and the 1-norm of the action vector must remain within the budget $B$.

\subsubsection{Heat Alerts Domain}\label{appdx:heat-alerts}

The heat alert issuance problem can be modeled as an MDP in the context of RL \cite{considine2023optimizing}. The state at any given time, denoted as  $\bs_t$, encompasses both exogenous and endogenous factors. Exogenous factors include current weather conditions, such as the heat index, temperature, and humidity, which directly influence the risk of heat-related health issues. Endogenous factors include the history of issued alerts, such as the number and timing of past alerts, their effectiveness, and the remaining budget for the season. Additionally, the day of the week is considered, as public responsiveness to alerts may vary between weekdays and weekends.
The action space is binary, with $\ba_t\in\mathbb{Z}_2$. The decision to issue a heatwave alert $\ba_t=1$ or not $\ba_t=0$ is constrained by the remaining alert budget. If the budget is exhausted, no further alerts can be issued. The reward function is designed to reflect the reduction in heat-related hospitalizations, which depends on the effectiveness of the alert under current weather conditions. A Bayesian hierarchical framework could be employed to model the health impact of alerts, capturing the uncertainty in their effectiveness. Importantly, consecutive alerts tend to lose effectiveness, introducing a diminishing returns effect that must be accounted for in the decision-making process.

The transition dynamics, 
$P(\bs_{t+1} | \bs_t, \ba_t)$, describe how the system evolves over time. The next state is influenced by weather trajectories, the action taken, and public responsiveness to alerts. For instance, issuing an alert reduces the remaining budget and updates the history of issued alerts, while the weather conditions may change independently. Public responsiveness may also vary based on the frequency and timing of past alerts.
A key constraint in this problem is the limited alert budget, which necessitates strategic allocation of alerts throughout the season. The goal is to learn a policy 
$\pi(\bs_t|\ba_t)$ that maximizes cumulative rewards by effectively issuing alerts during severe heat conditions to minimize hospitalizations, while conserving the budget for future use. This involves balancing immediate health benefits against the potential need for alerts later in the season, addressing the trade-offs between short-term and long-term outcomes.

For this domain, the budget constraints can be expressed as $\sum_{t=1}^h \ba_t\in\mathbb{R}\leq B$, where the total sum of all actions over time horizon $h$ must not exceed a budget threshold $B$.

\subsection{Gym environments an Language Wrappers}

We implemented the \texttt{WearableDevicesAssignment} environments as \text{gymnasium} environments \cite{towers2024gymnasium}, while the \texttt{HeatAlerts} domain was already available in this format. We additionally created a \texttt{LanguageWrapper} Python class described in Table \ref{tab:language_wrapper}, which can be applied to any \texttt{gymnasium} environment. Our code implementations can be applied to any environment wrapped in this class.

\begin{table}[h]
    \centering
    \scalebox{0.75}{
    \begin{tabular}{|l|l|p{12cm}|}
        \hline
        \textbf{Method/Property} & \textbf{Type} & \textbf{Description} \\
        \hline
        \texttt{task\_text} & Property (Abstract) & Returns a description of the task that the environment is solving. \\
        \hline
        \texttt{action\_space\_text} & Property (Abstract) & Returns a description of the action space of the environment. \\
        \hline
        \texttt{state\_descriptor(obs, info)} & Abstract Method & Converts the observation into a text description. \\
        \hline
        \texttt{step(action)} & Method & Wraps the step method of the environment adding the text representation to the state space. \\
        \hline
        \texttt{reset(seed, options)} & Method & Wraps the reset method of the environment adding the text representation to the state space. \\
        \hline
        \texttt{action\_parser(s)} & Method & Parses an action string and converts it into an appropriate format for the environment's action space. \\ \hline
        \texttt{(rule\_examples} & Property (Optional) & Returns a list of string representation of rules. \\
        \hline
    \end{tabular}
    }
    \caption{Methods and properties of the \texttt{LanguageWrapper} class}
    \label{tab:language_wrapper}
\end{table}

\subsection{RL implementations, hyperparameters and Settings}

We implemented three main RL algorithms for the experiment sections: Attention-based SAC for \rbrl, numeric PPO, and Finetuning-based PPO. We based our implementation on the single-file, high-quality implementations from the \texttt{cleanrl} project \citep{huang2022cleanrl}. For Attention-based SAC, we required significant changes to keep track of the rule-augmented state space, as described in Section \ref{sec:sac-rl}. Other major changes to the baseline SAC implementation (originally designed for Atari) were more frequent target network updates and updating the actor and critic four times per iteration. This was done to improve sample efficiency and cope with the slow generation by the LLM. Numeric PPO was used practically without modification.

For the Finetuning-based PPO, we used low-rank adaptation (LoRA) \cite{hu2021lora} with the Transformers package and models hosted on Llama Hugging Face \cite{wolf-etal-2020-transformers}. We set the rank to $r=1$ and the adaptation weight to 2, resulting in only 0.8\% trainable parameters (still an order of magnitude larger than the Attention-based policy). Tables \ref{tab:parameters:sac}, \ref{tab:parameters:nppo}, and \ref{tab:parameters:lora} show the hyperparameters and settings used in these implementations.

\subsection{Computing environment}

SAC attention can run on a regular laptop since most of the computation happens in the cloud through API LLM calls, while the RL module is small and can run on personal CPUs. Nonetheless, the process is bottlenecked by the speed of generation from the APIs. A full run of 2 million environment steps, with parallelized API calls across four environments, took approximately four hours to complete. One training cycle did not exceed \$10 in API costs. However, all the experiments and development incurred approximately \$1,500 in API costs.
As described in the main text, the LLM fine-tuning experiments used an Nvidia A100 40GB GPU for each seed, training on three seeds for 18 hours each. Computations were performed on a Slurm-based high-performance computing cluster.

\begin{table*}[!h]
\centering
\caption{SAC Hyperparameters and Settings for \rbrl.}
\scalebox{0.6}{
\label{tab:parameters:sac}
\begin{tabular}{lll}
\toprule
\textbf{Parameter} & \textbf{Default Value} & \textbf{Description} \\
\midrule
\texttt{num\_envs} & 4 & Number of parallel environments \\
\texttt{total\_timesteps} & 500 & Total number of environment steps \\
\texttt{gamma} & 0.95 & Discount factor \(\gamma\) \\
\texttt{tau} & 1.0 & Target smoothing coefficient \\
\texttt{batch\_size} & 16 & Batch size of sample from the replay memory \\
\texttt{buffer\_size} & 4096 & The replay memory buffer size \\
\texttt{max\_episode\_steps} & 32 & Episode truncation  \\
\texttt{learning\_starts} & 256 & Timestep to start learning \\
\texttt{policy\_lr} & \(1 \times 10^{-4}\) & Learning rate of policy network optimizer \\
\texttt{q\_lr} & \(1 \times 10^{-4}\) & Learning rate of Q-network optimizer \\
\texttt{actor\_updates} & 4 & Number of actor updates per update cycle \\
\texttt{critic\_updates} & 4 & Number of critic updates per update cycle \\   
\texttt{target\_network\_frequency} & 64 & The frequency for the target network update\\
\texttt{alpha} & 0.01 & Initial entropy regularization coefficient \\
\texttt{autotune} & True & Automatic tuning of the entropy coefficient \\
\texttt{target\_entropy\_scale} & 0.89 & Coefficient for scaling the autotune entropy target \\ 
\texttt{dropout} & 0.05 & The dropout rate \\
\texttt{num\_rules} & 5 & Number of rules for \rbrl \\ 
\texttt{llm} & "gpt-4o-mini" & LLM for generation \\
\texttt{embedder\_lm} & "m2-bert-80M-8k-retrieval" & The LLM to use for embeddings \\
\texttt{embed\_dim} & 768 & Dimension of rule embeddings \\
\texttt{hidden\_dim} & 16 & Hidden dimension of networks \\
\texttt{rule\_reward\_coef} & 1.0 & The reward coefficient for the rules \\
\texttt{num\_self\_attention\_layers} & 1 & For the actor and critic\\
\texttt{num\_cross\_attention\_layers} & 1 & For the actor and critic\\
\bottomrule
\end{tabular}
}
\end{table*}

\begin{table}[!h]
\centering
\caption{Numeric PPO Hyperparameters and Settings.}
\label{tab:parameters:nppo}
\scalebox{0.6}{
\begin{tabular}{lll}
\toprule
\textbf{Parameter} & \textbf{Default Value} & \textbf{Description} \\
\toprule
\texttt{total\_timesteps} & 50000 & Total timesteps of the experiments \\
\texttt{learning\_rate} & \(2.5 \times 10^{-4}\) & Learning rate of the optimizer \\
\texttt{num\_envs} & 4 & Number of parallel environments \\
\texttt{num\_steps} & 512 & Steps per policy rollout \\
\texttt{anneal\_lr} & False & no learning rate annealing \\
\texttt{gamma} & 0.95 & Discount factor \(\gamma\) \\
\texttt{gae\_lambda} & 0.95 & Lambda for Generalized Advantage Estimation \\
\texttt{num\_minibatches} & 4 & Number of mini-batches \\
\texttt{update\_epochs} & 64 & Number of update epochs \\
\texttt{norm\_adv} & True & Whiten advantages \\
\texttt{clip\_coef} & 0.2 & Surrogate clipping coefficient \\
\texttt{clip\_vloss} & True & Clipped loss for value function \\
\texttt{ent\_coef} & 0.01 & Coefficient of entropy term \\
\texttt{vf\_coef} & 0.5 & Coefficient of value function \\
\texttt{max\_grad\_norm} & 0.5 & Maximum gradient clipping norm \\
\texttt{target\_kl} & None & Target KL divergence threshold \\
\texttt{hidden\_dim} & 16 & Hidden dimension of networks \\
\texttt{num\_hidden\_layers} & 2 & For policy and critic networks \\
\texttt{max\_episode\_steps} & 32 & Episode truncation  \\
\bottomrule
\end{tabular}
}
\end{table}

\begin{table}[!h]
\centering
\caption{LLM PPO Finetuning Hyperparameters and Settings.}
\label{tab:parameters:lora}
\scalebox{0.6}{
\begin{tabular}{lll}
\toprule
\textbf{Parameter} & \textbf{Default Value} & \textbf{Description} \\
\toprule
\texttt{total\_timesteps} & 500 & Total number of timesteps \\
\texttt{learning\_rate} & \(2.5 \times 10^{-4}\) & Learning rate of optimizer \\
\texttt{num\_envs} & 4 & Number of parallel game environments \\
\texttt{num\_steps} & 32 & Steps per policy rollout \\
\texttt{anneal\_lr} & True & Enable learning rate annealing \\
\texttt{gamma} & 0.95 & Discount factor \(\gamma\) \\
\texttt{gae\_lambda} & 0.95 & Lambda for Generalized Advantage Estimation \\
\texttt{update\_epochs} & 4 & Number of update epochs per cycle \\
\texttt{norm\_adv} & True & Advantages whitening \\
\texttt{clip\_coef} & 0.2 & Surrogate clipping coefficient \\
\texttt{clip\_vloss} & True & Clipped loss for value function \\
\texttt{ent\_coef} & 0.01 & Coefficient of entropy term \\
\texttt{vf\_coef} & 0.5 & Coefficient of value function \\
\texttt{kl\_coef} & 0.05 & KL divergence with reference model \\
\texttt{max\_grad\_norm} & 0.5 & Maximum gradient clipping norm \\
\texttt{target\_kl} & None & Target KL divergence threshold \\
\texttt{dropout} & 0.0 & Dropout rate \\
\texttt{llm} & "meta-llama/Llama-3.1-8B-Instruct" & Model to fine-tune \\
\texttt{train\_dtype} & "float16" & Training data type \\
\texttt{gradient\_accumulation\_steps} & 16 & Number of gradient accumulation steps \\
\texttt{minibatch\_size} & 1 & Mini-batch size for fine-tuning \\
\texttt{max\_chunk\_size} & 256 & Maximum length sequence for the back propagation \\
\texttt{max\_episode\_steps} & 32 & Maximum number of steps per episode \\
\bottomrule
\end{tabular}
}
\end{table}

\newpage

\section{Additional Survey Results}
\begin{figure}[H]
\centering
\begin{minipage}{.4\textwidth}
\centering
\includegraphics[width=1\columnwidth]{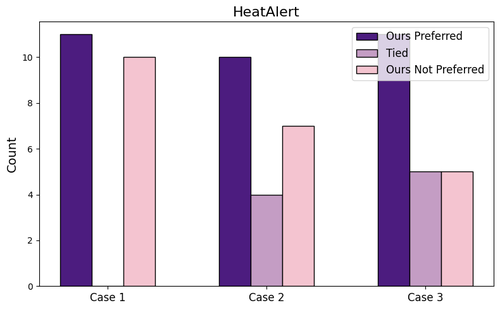}
\subcaption{\texttt{HeatAlert}}
 \label{fig:heatalert_human}
\end{minipage}\hfill
\begin{minipage}{.4\textwidth}
\centering
\includegraphics[width=1\columnwidth]{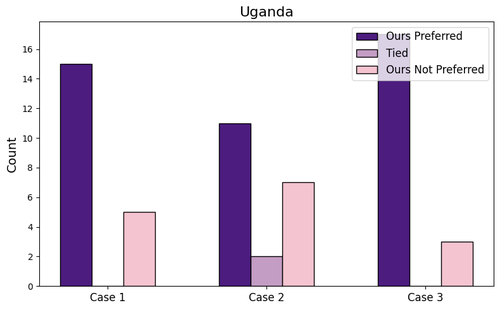}
\subcaption{\texttt{Uganda}}
 \label{fig:Uganda_human}
\end{minipage}
\caption{Results from \textbf{human} surveys conducted in the \texttt{HeatAlert} (a) and \texttt{Uganda} (b) environments. 21 participants provided feedback for the \texttt{HeatAlert} domain, while 20 valid responses were collected for the \texttt{Uganda} domain. The results indicate that our method (``Ours Preferred") was favored by a majority of participants, particularly in the \texttt{Uganda} domain, where the preference is more pronounced.}
        \label{fig:survey_humans}
\end{figure}

Figure \ref{fig:survey_humans} illustrates the results of a human survey conducted to evaluate the quality of explanations generated by our method compared to alternatives. A total of 21 valid responses were collected for the \texttt{HeatAlert} environment (Figure \ref{fig:heatalert_human}), and 20 valid responses were gathered for the \texttt{Uganda} environment (Figure \ref{fig:Uganda_human}).
As shown in the figures, our method was favored by the majority of participants across all cases. In the \texttt{HeatAlert} environment, the preference for our approach is evident, although there is a small percentage of tied and ``Not Preferred" responses. In contrast, the preference for our method is even more pronounced in the \texttt{Uganda} environment, with a significantly higher number of participants selecting ``Ours Preferred." These results demonstrate the effectiveness of our approach in generating explanations that resonate better with human users, particularly in the \texttt{Uganda} domain.

\begin{figure}[H]
\centering
\begin{minipage}{.4\textwidth}
\centering
\includegraphics[width=1\columnwidth]{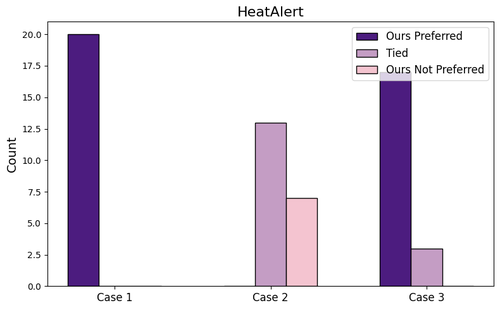}
\subcaption{\texttt{HeatAlert}}
 \label{fig:heatalert_LLM}
\end{minipage}\hfill
\begin{minipage}{.4\textwidth}
\centering
\includegraphics[width=1\columnwidth]{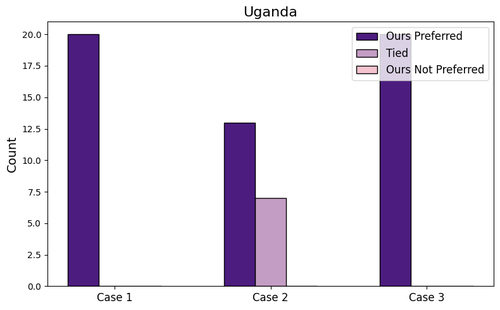}
\subcaption{\texttt{Uganda}}
 \label{fig:Uganda_LLM}
\end{minipage}
\caption{Survey results generated by querying \textbf{LLMs} 20 times for each case in the \texttt{HeatAlert} (a) and \texttt{Uganda} (b) environments. By varying the sampling temperature, randomized responses were collected for comparison. The results demonstrate that our method (``Ours Preferred") consistently outperforms alternatives across all cases, highlighting its robustness and alignment with the evaluation criteria of the LLMs.}
        \label{fig:survey_LLMs}
\end{figure}

Figure \ref{fig:survey_LLMs} illustrates the survey outcomes obtained by querying LLMs 20 times for each case in the \texttt{HeatAlerts} (Figure \ref{fig:heatalert_LLM}) and \texttt{Uganda} (Figure \ref{fig:Uganda_LLM}) environments. To ensure variability, the LLM's sampling temperature was controlled, enabling randomized responses for each trial. Similar to the human survey results, our method (``Ours Preferred") is overwhelmingly favored across all cases in both domains. Notably, the consistency of ``Ours Preferred" responses highlights the effectiveness of our approach in generating explanations that align well with the LLM's evaluation criteria, further validating the robustness of our methodology.

\begin{figure}[H]
\centering
\begin{minipage}{.6\textwidth}
\centering
\includegraphics[width=1\columnwidth]{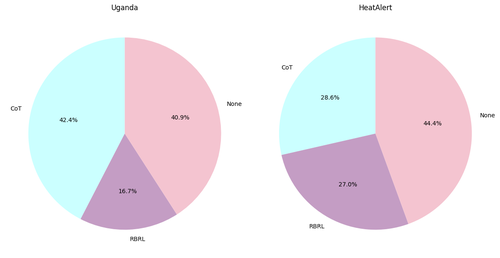}
\subcaption{Results from the \textbf{human} survey, showcasing the proportion of hallucination detected across three categories: \texttt{CoT}, \texttt{RBRL}, and \texttt{None}. In both domains, hallucinations were most frequently identified in \texttt{None}, with \texttt{RBRL} showing significantly fewer instances.}
 \label{fig:Hallucination_human}
\end{minipage}\hfill
\begin{minipage}{.6\textwidth}
\centering
\includegraphics[width=1\columnwidth]{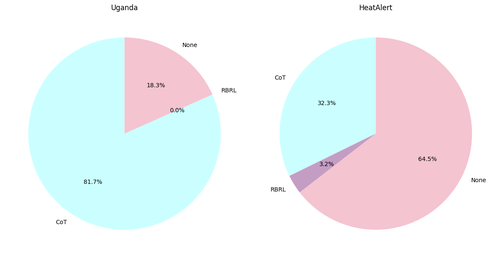}
\subcaption{Results from the \textbf{LLM}-based survey, where hallucination detection was assessed through multiple iterations of LLM evaluation. \texttt{CoT} exhibited higher hallucination rates in the \texttt{Uganda} domain, while \texttt{RBRL} demonstrated minimal hallucination occurrences in both domains.}
 \label{fig:Hallucination_LLM}
\end{minipage}
\caption{Survey results for hallucination detection across the \texttt{HeatAlert} and \texttt{Uganda} environments.}
        \label{fig:Hallucination_survey}
\end{figure}

Figure \ref{fig:Hallucination_survey} illustrates the survey results evaluating hallucination occurrences across two environments (\texttt{Uganda} and \texttt{HeatAlert}) for three explanation types: Chain of Thought (\texttt{CoT}), Rule-Bottleneck Reinforcement Learning (\texttt{RBRL}), and \texttt{None} (indicating no explanation).

In Figure \ref{fig:Hallucination_human}, the results from the human survey indicate that \texttt{CoT}-based explanations had a significant proportion of hallucinations, particularly in the \texttt{Uganda} environment, where it accounted for 42.4\% of responses. \texttt{RBRL} explanations showed markedly fewer hallucinations in both domains, highlighting its robustness. A notable percentage of responses for \texttt{None} indicate scenarios where explanations were either absent or irrelevant.
In Figure \ref{fig:Hallucination_LLM}, results from the LLM survey further emphasize the trends observed in the human survey. Hallucination rates for \texttt{CoT} were even higher in the \texttt{Uganda} environment (81.7\%), whereas \texttt{RBRL} explanations exhibited almost no hallucinations across both domains. In the \texttt{HeatAlert} environment, the absence of explanations (\texttt{None}) led to the highest percentage of hallucinations, underlining the importance of well-structured, rule-based explanations like \texttt{RBRL}.
These results collectively demonstrate that the \texttt{RBRL} framework significantly mitigates hallucinations, providing more accurate and reliable explanations compared to other methods.

\section{Prompt Templates and Rule Examples}\label{sec:prompt_appendix}

\subsection{Prompt Format}
\begin{figure}[H]
    \centering    \includegraphics[width=0.65\linewidth]{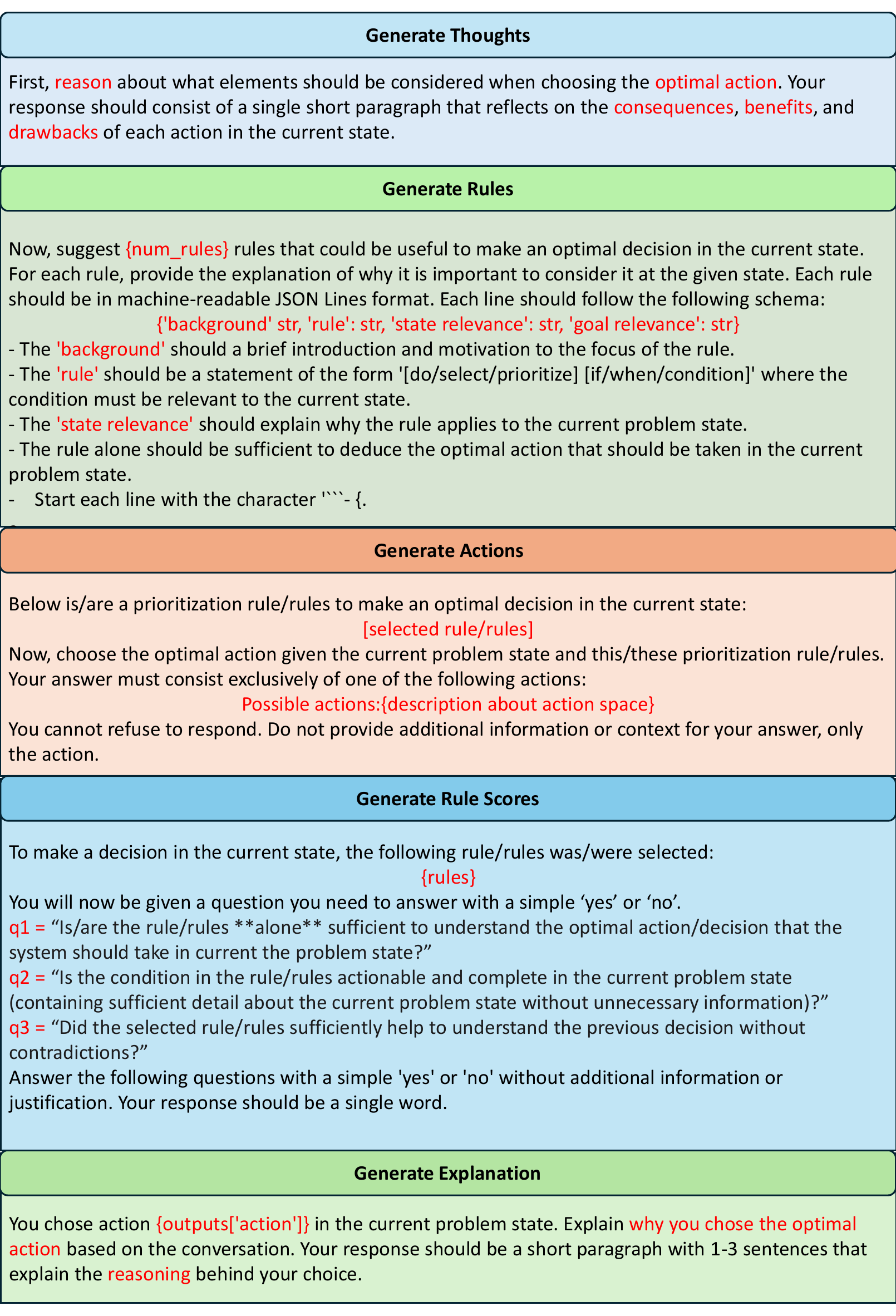}
    \caption{Prompts template for generating thoughts, rules, actions, rule scores, and explanations. }
    \label{fig:combined_prompt}
\end{figure}

In this section, we illustrate the prompt format used in our \rbrl for generating thoughts, rules, actions, rule scores, and explanations in Figure \ref{fig:combined_prompt}.

\subsection{Rule Example}
\begin{figure}[H]
    \centering
    \includegraphics[width=0.75\linewidth]{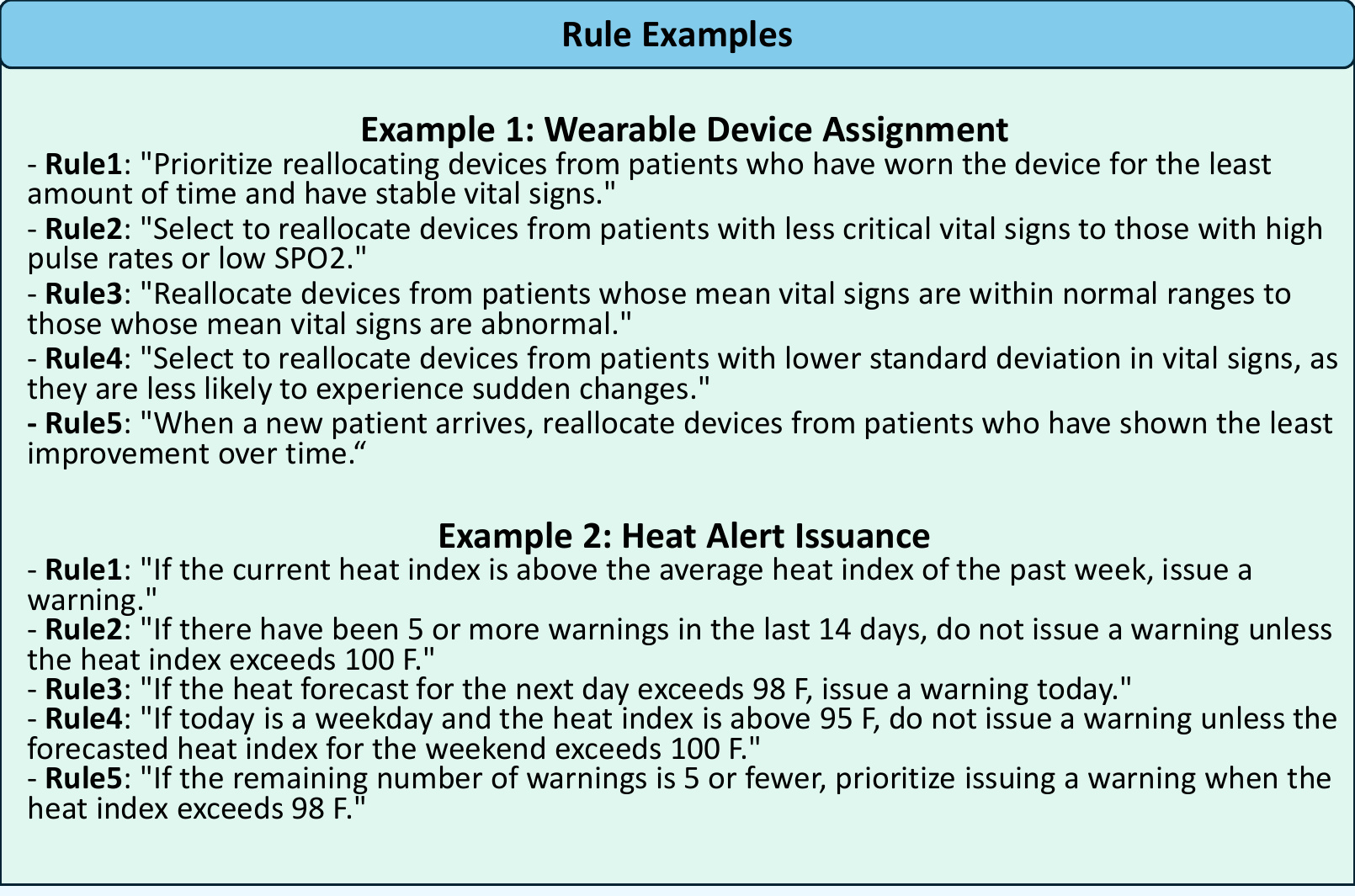}
    \caption{Rule examples for the considered two domains.}
    \label{fig:rule_example}
\end{figure}

In this section, we provide some rules for each domain in Figure \ref{fig:rule_example}.

\section{Survey Example}
\label{appdx:survey}
In this section, we present a survey example from the \texttt{Wearable\_Device\_Assignment} domain. The survey for the \texttt{HeatAlert} domain follows the same format and can be easily adapted by substituting the task and corresponding actions. For brevity, we include only one example case from the \texttt{Wearable\_Device\_Assignment}  domain.

\begin{figure}[H]
    \centering
    \includegraphics[width=0.7\linewidth]{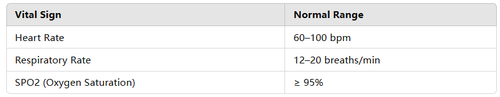}
    \caption{Normal range of vital signs.}
    \label{fig:vital sign}
\end{figure}

\textbf{Task:} \emph{You are tasked with optimizing the allocation of limited vital sign monitoring devices among patients. Devices improve vital signs and prevent abnormalities, but their limited availability requires reallocating them from stable patients to higher-risk incoming patients, who must always receive a device. The normal range of vital signs are provided in Figure \ref{fig:vital sign}. The goal is to minimize costs associated with abnormal vital signs, where costs are calculated exponentially based on deviations from predefined thresholds. Wearing a device improves abnormal vital signs with a 70\% success rate.}

\textbf{Possible actions:} \emph{Choose the id of the device that will be reallocated to the new incoming patient. Your answer should be a single integer i from 0 to 4 (the number of devices) such that}:
\begin{itemize}
    \item \emph{Always choose a free device if available}
    \item \emph{If no free device is available, then choose device i whose current patient is at least risk or would benefit less from wearing the device.}
\end{itemize}

\emph{In the following, you will be presented with three cases. Each case includes two explanations. Please read the text for each case carefully and answer the questions provided.}

\textbf{Case 1:}
Current state of the decision problem: \\
Number of devices: 5 \\
Number of free devices: 1\\
IDs of free devices: 3

\textbf{Device 0:} Device is currently assigned to a patient with the following description: \\
*Timesteps wearing the device*: 1\\ 
*Pulse rate* - Last value: 95.22 - Mean: 105.12 - Standard deviation/volatility: 10.56 \\
*Respiratory rate* - Last value: 20.14 - Mean: 20.54 - Standard deviation/volatility: 0.64\\
 *SPO2* - Last value: 98.42 - Mean: 97.89 - Standard deviation/volatility: 0.88

\textbf{Device 1:} Device is currently assigned to a patient with the following description:\\ 
*Timesteps wearing the device*: 1\\ 
*Pulse rate* - Last value: 89.55 - Mean: 81.86 - Standard deviation/volatility: 8.55 \\
*Respiratory rate* - Last value: 14.85 - Mean: 20.81 - Standard deviation/volatility: 3.40 \\
*SPO2* - Last value: 95.31 - Mean: 96.22 - Standard deviation/volatility: 1.38 

\textbf{Device 2:} Device is currently assigned to a patient with the following description: \\
*Timesteps wearing the device*: 1 \\
*Pulse rate* - Last value: 106.05 - Mean: 105.09 - Standard deviation/volatility: 2.91 \\
*Respiratory rate* - Last value: 19.34 - Mean: 20.80 - Standard deviation/volatility: 2.69 \\
*SPO2* - Last value: 99.56 - Mean: 99.36 - Standard deviation/volatility: 0.27 

\textbf{Device 3:} Device is currently free. 

\textbf{Device 4:} Device is currently assigned to a patient with the following description: \\
*Timesteps wearing the device*: 1 \\
*Pulse rate* - Last value: 80.02 - Mean: 79.03 - Standard deviation/volatility: 1.58 \\
*Respiratory rate* - Last value: 22.71 - Mean: 21.31 - Standard deviation/volatility: 5.45 \\
*SPO2* - Last value: 99.61 - Mean: 99.86 - Standard deviation/volatility: 0.14

\underline{\textbf{Explanation A:}} I chose to reallocate device 4 because the patient currently using it has a stable pulse rate (80.02) and a high SPO2 level (99.61), indicating they are less at risk and may not require continuous monitoring. In contrast, reallocating this device allows for an incoming patient, who likely has more urgent health needs, to receive the device, thereby optimizing the overall allocation of resources to those who require immediate attention.

\underline{\textbf{Explanation B:}} I chose action {`device': 3} because device 3 is currently free and must be assigned to an incoming patient who requires immediate monitoring to prevent any potential deterioration in their health. This decision aligns with the prioritization rule that emphasizes the importance of providing care to all patients, especially those at risk, ensuring that the incoming patient receives the necessary vital sign monitoring.

\textbf{Q1.  Do Explanation A and Explanation B appear the same or different to you?}

\noindent
$\square$ Same (Skip Q2 and go to Question Q3) \\
$\square$ Different \\

\textbf{Q2. Which explanation do you find better?}

\noindent
$\square$ Explanation A \\
$\square$ Explanation B \\

\textbf{Q3. Do the explanations contain any hallucinations?}

\noindent
$\square$ Both \\
$\square$ Only Explanation A \\
$\square$ Only Explanation B \\
$\square$ None \\

\end{document}